\documentclass[11pt]{article}
\usepackage[utf8]{inputenc}

\usepackage{epsfig,amsmath,latexsym,amssymb}
\usepackage{graphicx}
\usepackage{lscape}
\usepackage{picture, eso-pic, tikz} 
\usepackage{dsfont}
\usepackage{appendix}
\usepackage{multirow}
\usepackage{bbm}
\usepackage{listings}
\usepackage{adjustbox}

\usepackage{rotating}
\usepackage{mdframed}
\usepackage{etoolbox}
\BeforeBeginEnvironment{appendices}{\clearpage}

\def\R{{\mathbb R}}  
\def\N{{\mathbb N}}  

\DeclareMathOperator*{\argmin}{arg\,min}

\newcommand{\Remm}[1]{}

\newtheorem{model ass}[theo]{Model Assumptions}

\numberwithin{equation}{section}

\definecolor{MyGray}{rgb}{0.92,0.92,0.92}


\def\bx{\boldsymbol{x}}
\def\by{\boldsymbol{y}}
\def\bz{\boldsymbol{z}}

\begin{document}

	\author{
Ronald Richman\footnote{Old Mutual Insure and University of the Witwatersrand, Johannesburg, South Africa; ronaldrichman@gmail.com} \and
	Salvatore Scognamiglio \footnote{Department of Management and Quantitative Studies, University of Naples ``Parthenope",\newline salvatore.scognamiglio@uniparthenope.it}
}

\date{Version of \today}
\title{Multiple Yield Curve Modeling and Forecasting using Deep Learning}
\maketitle

\begin{abstract}
	\noindent	
This manuscript introduces deep learning models that simultaneously describe the dynamics of several yield curves. We aim to learn the dependence structure among the different yield curves induced by the globalization of financial markets and exploit it to produce more accurate forecasts. By combining the self-attention mechanism and nonparametric quantile regression, our model generates both point and interval forecasts of future yields. The architecture is designed to avoid quantile crossing issues affecting multiple quantile regression models. Numerical experiments conducted on two different datasets confirm the effectiveness of our approach. Finally, we explore potential extensions and enhancements by incorporating deep ensemble methods and transfer learning mechanisms.

~

\end{abstract}

\section{Introduction}

Yield curves are used for a wide variety of tasks in actuarial science and finance for deriving the present value of future cashflows within valuations that apply a market consistent approach. A market consistent approach is required by modern solvency regulations, such as Solvency II, while recently updated accounting standards, such as the recently introduced IFRS 17, require the use of credit and liquidity adjusted yield curves for discounting liabilities, including both life and non-life insurance liabilities. Insurers, and other entities, that report on their liabilities on a discounted basis are exposed to the risk of changes in the interest rates in their markets, which translate directly into changes in the solvency of these entities. Therefore, managing this risk of adverse changes in yield curves - which we refer to as interest rate risk in what follows - is an important task within actuarial work, which is usually considered in the context of corresponding changes in the asset portfolio backing these liabilities, changes in the value of which may act as an offset. This process is, therefore, usually referred to as Asset-Liability Management (ALM). Moreover, insurers are required to hold capital to ensure that their solvency is adequately protected in most solvency regimes, such as Solvency II. To measure the extent of the interest rate risk, as well as the corresponding capital needed to be held, insurers and other financial institutions often rely on modelling the uncertain future evolution of the yield curve using a variety of different models. Once the yield curves have been modelled, the models are used to derive scenarios for the future evolution of the yield curves, which are then applied to derive capital requirements. Here, we distinguish between unconditional and conditional approaches to yield curve modelling: the former approach calibrates models of the yield curve evolution using historical data at a point in time, derives stresses based on these, and then applies these stresses without recalibrating these based on current market conditions. This approach underlies, for example, the standard formula approach of the Solvency II regulation. On the other hand, the conditional approach uses current market information to recalibrate yield curve stresses; this approach is often taken in internal model approaches within the Solvency II regulation.

Modelling interest rate risks is made more difficult due to the complexity of requirements of recent accounting standards, as well as the interconnected nature of financial markets across asset classes and geographies. The recent IFRS 17 standard departs from a purely market consistent valuation approach by requiring insurers to use yield curves that are modified to correspond to the financial characteristics of the liabilities being valued, as well as the asset portfolios backing these. While we do not explain this in detail, in short, insurers must derive yield curves consisting of the (credit) risk-free interest rate, as well as an allowance for an illiquidity premium. Thus, insurers reporting under IFRS 17 calibrate several different yield curves for discounting liabilities, the evolution of which will differ depending on both how risk-free rates and illiquidity premia change over time. Another reason that insurers may need to model the evolution of multiple yield curves is due to their investing in assets with different levels of credit quality; to manage the risk of the asset portfolio, it is often necessary to calibrate multiple yield curves which take account of credit-risk premia and model the (joint) evolution of these. Finally, insurers operating in multiple geographic jurisdictions need to manage the interest rate risk arising from changes in the different yield curves used in these markets. In all of these scenarios, it is not sufficient merely to model the dynamics of each yield curve on an independent basis, since this approach will not capture the correlation between asset classes and geographies and may lead to misstated estimates of risk and capital; rather the joint future evolution of the complete set of yield curves used by the insurer must be modelled. 

In this work, we focus on exactly this problem of jointly modelling and forecasting multiple yield curves for interest rate risk management, ALM and derivation of capital requirements; we note that this is done on the real-world basis and not the risk-neutral basis which is useful for option valuation. For this task, we use neural network models trained jointly on a significant amount of historical yield curve data across geographies to forecast yield curves on an expected (best-estimate) basis, as well as to forecast the quantiles of the yield curves; the latter can be used directly for risk management purposes, for example, calculating the Value-at-Risk of the insurer.

\subsection*{Literature review}

Several different approaches have appeared in the literature to model the uncertain future evolution of the yield curve \cite{teichmann2016consistent, venter2004testing}.

One class of models - focused on the risk-neutral evolution of the yield curve - consists of arbitrage-free models, which impose restrictions on the evolution of the yield curve to avoid risk-free profit opportunities. Prominent examples are the models developed in \cite{hull1990pricing} and \cite{heath1992bond}. Although these models are widely used in option pricing, without further adaptation, they are often found to forecast poorly compared with a simple random walk model (see \cite{duffee2002term}). We refer to \cite{yasuoka2018interest}, and the citations therein, for interesting discussions of adapting risk-neutral interest rate models for real-world purposes by estimating the market-price of interest rate risk. A popular commercial approach used by some insurers for internal modelling of interest rate (and other market) risks in the Solvency II capital regime consists of modifying arbitrage-free models to ensure that the implied future evolution of yield curves is constrained to meet certain real-world economic and market-variable targets, however, there is relatively little discussion of this in the academic literature.

A well-known technique is to apply principal components analysis (PCA) to vectors of the changes in the yield curve at each term and then to use the simulated changes in the yield curve to derive a distribution of yield curves (for an overview, see \cite{Redfern2014}). This approach was used, for example, to calibrate the yield curve stresses in the interest-rate risk module of Solvency II. This is an example of the unconditional approach to yield curve modelling, since the PCA analysis was performed at a point in time in the past, and is assumed to still be relevant in current market conditions.

Other authors follow a purely statistical approach. This class of models has evolved from univariate \cite{fama1987information} to multivariate time series models and recent advances in dynamic factor models. Among them, the dynamic extension of the well-known Nelson–Siegel (NS) model \cite{nelson1987parsimonious} proposed in \cite{diebold2006forecasting} (from now on referred to as DNS) has become very popular among practitioners thanks to its simplicity and discrete forecasting ability. In addition to being a relatively simple and parsimonious model, the DNS approach is also appealing since the NS model (and its extensions) underlying the DNS approach is often used by central banks and other institutions for calibrating the yield curve.  
Other notable examples of factor models can be found in \cite{bowsher2008dynamics, hardle2016yield}.


Numerous extensions of the DNS approach have been developed in the literature. Some resarch investigated using more flexible versions of the Nelson-Siegel model, for example, the model proposed in \cite{bliss1996testing}, the four-factor extension suggested in \cite{svensson1994estimating} and the five-factor model investigated in \cite{de2013modeling}. 
Other authors try to improve the forecasting performance of the DNS model, including some macroeconomics variables in the models (see \cite{diebold2006macroeconomy}). A nice overview of the NS and DNS models is in \cite{diebold2013yield}, who also provide economic intuitions for the factors used in the NS model. 

\subsubsection*{Multiple yield curve modeling}

Globalization has intensified the connection among the financial markets, inducing a dependence structure among different yield curves, which renders the process of modelling these jointly complex; moreover, above we have discussed other reasons for the need for joint modelling of multiple yield curves, which is a challenging task. Despite the relevance of the topic for financial markets, there is relatively little discussion of this topic in the literature on a real-world basis; on the other hand, more literature is available in the risk-neutral setting, see  \cite{cuchiero2016general} for an overview. Within the real-work setting, here we mention \cite{gerhart2020empirical}, who introduced a multiple-curves PCA method where the dynamics of multiple yield curves captured through Principal Component Analysis (PCA) are modelled as autoregressive processes and \cite{atkins2023improved}, who proposed a two-step method to jointly capture the risk-factor relationships within each curve and the risk-factor relationships between the curves. In a first stage, the authors use the PCA to derives components describing the dynamics of each curve, and then, secondly, combine these to describe the dynamics across all the curves. Notably, the joint forecasting of best-estimates and quantiles is not done in these works, and this is a novel aspect of the model presented here. 

\subsubsection*{Neural networks for yield curve modeling}

Recently, deep learning models have become popular for general machine learning tasks, due to their ability to model massive volumes of data in a flexible manner, and within finance and actuarial science.  Deep neural networks have been successfully applied to several tasks such as pricing \cite{barigou2022pricing, noll2020case}, reserving \cite{gabrielli2020neural} and mortality forecasting \cite{perla2021time, scognamiglio2022calibrating}. A detailed overview of the application of Artificial Intelligence (AI) and Machine Learning (ML) techniques in actuarial science can be found in \cite{richman2021ai1, richman2021ai2}. Focusing on yield curves modeling and interest rate risk management, the literature is relatively sparse. \cite{aljinovic2013neural} shows that feed-forward NN can be used to replace time series models to extrapolate the future values of the NS parameters. \cite{kauffmann2022learning} proposes to improve the flexibility of the DNS model using NN for deriving the factor loadings. \cite{nunes2019comparison} directly employ feed-forward NN to forecast future yields, and \cite{gerhart2019robust} use modern recurrent neural networks such as the Long Short-Term Memory which are specifically designed to analyse  sequential data, such as the time-series of the yields.

\subsubsection*{Contributions}

In this work, we develop deep learning models that simultaneously model and forecast the dynamics of the multiple yield curves, which could be related to different countries, credit qualities or liquidity characteristics; here, we focus on the first two of these. With their ability to describe high-dimensional time-series data and model the non-linearity often present in the data, deep learning techniques are promising tools for multiple yield curve modeling and forecasting. The idea is to exploit the dependence structure among the different yield curves induced by the globalization of financial markets or relationships between asset classes to improve the forecasting performance of our models by jointly training these on historical datasets of yield curves. Importantly, we focus both on producing best-estimate forecasts of the yield curve, as we as employ deep learning techniques to quantify the uncertainty around the predictions by forecasting quantiles. Although this latter aspect is relevant both from a practical risk management and a theoretical point of view, it has not been deeply investigated in the literature, and creating joint forecasts of these is, to our knowledge, novel. For uncertainty quantification we use modern deep learning approaches, such as non-parametric quantile regression and deep ensembles model \cite{lakshminarayanan2017simple}. This augmentation of best-estimate forecast models with forecast quantiles is particularly useful for risk management purposes since the quantiles correspond to the VaR, which, in practice, underlies many quantitative risk management systems in practice. A comparison among these methods for measuring the uncertainty in the forecasts is also of interest, since it could provide additional insights into how well these methods describe the evolution of the yield curves.

We utilize recent advances in deep learning methodology into our selected yield curve model, specifically, the self-attention mechanism, which has been used to great success in natural language processing \cite{vaswani2017attention} and has recently been applied for severity modeling of flood insurance claims \cite{kuo2021embeddings}. Here, we show how the features derived using a convolutional neural network can be enhanced using the self-attention mechanism for greater forecasting accuracy.

Finally, we investigate the use of transfer learning methods, which aim to transfer as much knowledge as possible from an existing model to a new model designed for a similar task. These methods are already intensively used in computer vision and natural language processing tasks, where models are trained on large general datasets, then fine-tuned on more specific tasks. In our context, the transfer learning mechanism is applied to exploit knowledge learned by parameterizing models on a database of yield curves from multiple jurisdictions, then transferring the learned model to a smaller dataset of yield curves derived for assets of various credit quality. \\

\textbf{Organization of the manuscript:} The rest of the manuscript is structured as follows. Section \ref{sec:DNS} introduces two of the most popular factor models for yield curve modeling and forecasting, Section \ref{sec:nn} describes neural network building blocks used in Section \ref{sec:core}, where we present the proposed yield curve model. Section \ref{sec:numerical} illustrates some numerical experiments on a large cross-geography dataset of yield curves, Section \ref{sec:extensions} discusses some possible ways to extend and enhance the proposed model, and Section \ref{sec:conclusions} concludes.

\section{Dynamic Nelson-Siegel and Nelson-Siegel-Svenson Models}
\label{sec:DNS}

We consider a scenario where the objective is to model the  dynamics of yield curves belonging to different families. These curve families might pertain to yield curves that vary in credit rating quality; for example, families could be labeled as {\tt `A'}, {\tt `AA'}, {\tt `AAA'}  and so forth. Alternatively, the yield curves could be associated with government bonds from different countries, with families labeled as {\tt `Euro'}, {\tt `United Kingdom'} and so on.

Let $\mathcal{I}$ represent the set of considered curve families, $\mathcal{M}$ denote the set of time-to-maturities for which the curve is defined, and $y_t^{(i)}(\tau)$ denote the continuously compounded zero-coupon nominal yield at time $t \in \mathcal{T}$ on a $\tau$-month bond (i.e. at tenor $\tau \in \mathcal{M}$) for the $i$-th bond in the set $\mathcal{I}$. Importantly, we note that here we work with spot rates, whereas, for example, PCA analysis of yield curves is often performed on forward rates. Here we focus on describing more traditional models which will be used as a benchmark for the neural network models introduced later.

In their influential work, Nelson and Siegel (NS) \cite{nelson1987parsimonious} introduce a three-factor model that, at a given date $t$, describes the relationship between the yield and maturity $\tau$. 
Given that the classical NS model is static, \cite{diebold2006forecasting} introduces a dynamic version where the model's parameters can vary over time. In this case, $y_t^{(i)}(\tau)$ can be expressed as follows:
\[
    y^{(i)}_{t}(\tau) = \beta^{(i)}_{0,t} + \beta^{(i)}_{1,t} \Big(\frac{1-e^{-\lambda^{(i)}_{t}\tau}}{\lambda^{(i)}_{t}\tau}\Big)+ \beta^{(i)}_{2,t} \Big( \frac{1-e^{-\lambda^{(i)}_{t} \tau}}{\lambda^{(i)}_{t} \tau}- e^{-\lambda^{(i)}_{t} \tau} \Big) + \epsilon_{t}^{(i)}(\tau),  \quad \epsilon_{t}^{(i)}(\tau) \sim \mathcal{N}(0, \sigma_{\epsilon^{(i)}}^2),
\]
where $\beta_{0,t}^{(i)},\beta_{1,t}^{(i)}, \beta_{2,t}^{(i)}, \lambda_t^{(i)} \in \mathbb{R}$ are model parameters governing the shape of the curve that are estimated for each date $t$ and each family curve $i$ by using market data.  More specifically, the parameters $\beta_{j,t}^{(i)}$ with $j \in \{0, 1, 2\}$ can be interpreted as three latent factor factors defining the level, slope and curvature of the yield curve, respectively, while $\lambda_t^{(i)}$ indicates where the loading achieves its maximum; here, we follow the interpretation of these factors given in \cite{diebold2006forecasting}.
The calibration of the NS model with respect to all the parameters raises an optimization problem that is intrinsically nonlinear due to the $\lambda_t^{(i)}$ parameter.  However, since it doesn't impact the results, many authors, including \cite{diebold2006forecasting}, suggest keeping $\lambda_t^{(i)} = \lambda^{(i)}, t \in \mathcal{T}$ fixed, and estimating the remaining parameters by solving, for each family $i$, the sequence of linear optimization problems: 

\[
\argmin_{\beta_{0,t}^{(i)},\beta_{1,t}^{(i)}, \beta_{2,t}^{(i)}}  \sum_{\tau \in \mathcal{M}} \bigg(   y^{(i)}_{t}(\tau) -  \beta^{(i)}_{0,t} -\beta^{(i)}_{1,t} \Big(\frac{1-e^{-\lambda_{i}\tau}}{\lambda^{(i)}_{t}\tau}\Big) - \beta^{(i)}_{2,t} \Big( \frac{1-e^{-\lambda^{(i)}_{t} \tau}}{\lambda^{(i)}_{t} \tau}- e^{-\lambda^{(i)}_{t} \tau} \Big)  \bigg)^2, \forall t \in \mathcal{T}.
\]

To make forecasts,  a dynamic model for the latent factors $\boldsymbol{\beta}_{j,t}^{(i)}$ with $j \in \{0, 1, 2\}$  has to be specified. The simplest choice consists of using a set of individual first-order Autoregressive (AR) models: 

\[
    {\beta^{(i)}_{j,t}} = \psi{0,j}^{(i)} + \psi_{1,j}^{(i)} {\beta^{(i)}_{j,t-1}} +\zeta_{t}^{(i)},  \quad \zeta_{t}^{(i)} \sim \mathcal{N}(0, \sigma_{\zeta^{(i)}}^2)
\]

where $\psi_{0,j}^{(i)}, \psi_{1,j}^{(i)} \in \mathbb{R}, i \in \mathcal{I}, j \in \{0, 1, 2\}$ are  the time-series model parameters and $\zeta_{t}^{(i)}$ are normally distributed error terms. Alternatively,  one could also model the vector $\boldsymbol{\beta}_t^{(i)} = \big(\beta_{0,t}^{(i)}, \beta_{1,t}^{(i)}, \beta_{2,t}^{(i)}\big) \in \mathbb{R}^3$ using single first-order multivariate Vector Autoregressive (VAR) model:

\[
    {\boldsymbol{\beta}}_t^{(i)} = \boldsymbol{a}^{(i)}_0 + A^{(i)} {\boldsymbol{\beta}}_{t-1}^{(i)}+ \boldsymbol{\eta}^{(i)}_t, \quad \quad \quad \quad\boldsymbol{\eta}^{(i)}_t \sim \mathcal{N}(0, E^{(i)})
\]

with $\boldsymbol{a}^{(i)}_0 \in \mathbb{R}^3, A^{(i)} \in \mathbb{R}^{3 \times 3}$, and $\boldsymbol{\eta}^{(i)}_t \sim N(0, E^{(i)})$ is the normally distributed error term with matrix $E^{(i)} \in \mathbb{R}^{3 \times 3}$.

Numerous extensions of the NS model and its dynamic version have been proposed in the literature. One of the most popular enhancements, due to Svensson \cite{svensson1994estimating}, introduces an additional term to augment flexibility. This extension enhances the model's capacity to capture various shapes of yield curves by incorporating additional curvature components. The Svensson extension allows for a more general representation of the term structure of interest rates, establishing it as a valuable and often used tool in fixed-income and financial modeling. The Nelson-Siegel-Svensson (NSS) model is defined as follows:

\[
    y^{{(i)}}_{{t}}(\tau) =\beta^{{(i)}}_{0,{t}} + \beta^{{(i)}}_{1,{t}} \Big(\frac{1-e^{-\lambda_{1,t}^{(i)}\tau}}{\lambda_{1,t}^{(i)}\tau}\Big)+ \beta^{{(i)}}_{2,{t}}\Big( \frac{1-e^{-\lambda_{1,t}^{(i)} \tau}}{\lambda_{1,t}^{(i)} \tau}- e^{-\lambda_{1,t}^{(i)} \tau} \Big) + \beta^{{(i)}}_{3,{t}}\Big( \frac{1-e^{-\lambda_{2,t}^{(i)} \tau}}{\lambda_{2,t}^{(i)} \tau}- e^{-\lambda_{2,t}^{(i)}\tau} \Big)+ \epsilon_{t}^{(i)}(\tau),
\]

where $\beta^{{(i)}}_{3,{t}} \in \mathbb{R}$ is a second curvature parameter, and $\lambda_{1,t}^{(i)},\lambda_{2,t}^{(i)}\in \mathbb{R}$ are two decay factors. Forecasts are obtained by applying the procedure adopted by \cite{diebold2006forecasting}.

\section{Neural Networks}
\label{sec:nn}

Neural networks (NN) represent nonlinear statistical models originally inspired by the functioning of the human brain, as implied by their name, and subsequently extensively developed for multiple applications in machine learning, see \cite{goodfellow2016deep} for a review. A feed-forward neural network comprises interconnected computational units, or neurons, organized in multiple layers. These neurons ``learn'' from data through training algorithms. The fundamental concept involves mapping input data to a new multi-dimensional space, extracting derived features. The output (target) is then modeled as a nonlinear function of these derived features; this process is called representation learning \cite{bengio2013rep}.  Implementing multiple feed-forward network layers is called deep learning in the literature, and has proved to be particularly promising when dealing with high-dimensional problems requiring the identification of nonlinear dependencies. The arrangement of connections among the units delineates various types of neural networks. Our neural network model is constructed on the principles of both feed-forward and recurrent neural networks. An overview of the neural network blocks employed in the best performing model presented in this paper is provided below, whereas we summarize briefly the network blocks that are used in models that perform less well. For more detail on neural networks in an actuarial context, we refer to \cite{Wuthrich2021}, whose notation we follow.

\subsection{Fully-Connected Layer}
A fully connected network (FCN) layer, also commonly known as a dense layer due to the dense connections between units, is a type of layer in a neural network where each neuron or unit is connected to every neuron in the previous layer. In other words, each neuron in a fully connected layer receives input from all the neurons in the preceding layer; if the layer is the first in a network, then each neuron is receives inputs from all of the covariates input into the network.

Let $\boldsymbol{x} \in \mathbb{R}^{q_0}$ be the input vector; a FCN layer with $q_1 \in \mathbb{N}$ units is a vector function that maps $\boldsymbol{x}$ into a $q_1$-dimensional real-valued space: 

\begin{equation*}\label{FCN layer}
	\boldsymbol{z}^{(1)} : \mathbb{R}^{q_0} \to \mathbb{R}^{q_1}, \quad \quad \boldsymbol{x} \mapsto  \boldsymbol{z}^{(1)} (\boldsymbol{x}) = \left( z^{(1)} _1(\boldsymbol{x}), z^{(1)} _2 (\boldsymbol{x}), \dots , z^{(1)} _{q_1}(\boldsymbol{x}) \right)'. 
\end{equation*}

The output of each unit is a new feature $z^{(1)} _j(\boldsymbol{x})$, which is a non-linear function of $\boldsymbol{x}$: 

\begin{equation*}
z^{(1)}_j(\boldsymbol{x})   = \phi \bigg( w^{(1)} _{j,0} + \sum_{l = 1}^{q_1} w^{(1)}_{j,l} x_l \bigg)  \quad \quad j = 1, 2, ..., q_1 , 
\end{equation*}
where $\phi : \mathbb{R} \mapsto \mathbb{R}$ is the activation function and  $w^{(1)}_{j,l}\in \mathbb{R}$ represent the weights.
 In matrix form, the output $\boldsymbol{z}^{(1)} (\boldsymbol{x})$ of the FCN layer can be written as:
  \begin{equation}
\boldsymbol{z}^{(1)} (\boldsymbol{x}) = \phi \big( \boldsymbol{w}^{(1)}_0 +  W^{(1)} \boldsymbol{x} \big).
\end{equation}

Shallow neural networks are those networks with a single dense layer and directly use the features derived in the layer for computing the (output) quantity of interest $y \in \mathcal{Y}$. In the case of  $\mathcal{Y} \subseteq \mathbb{R}$, the output of shallow NN reads: 

\begin{equation*}
y = \phi\left(w^{(o)}_0 + \langle \boldsymbol{w}^{(o)}, \boldsymbol{z}^{(1)} (\boldsymbol{x}) \rangle \right), 
\end{equation*}

where $w^{(o)}_0 \in \mathbb{R}$, $\boldsymbol{w}^{(o)} \in \mathbb{R}^{q_1}$, $\langle \cdot, \cdot \rangle$ denotes the scalar product in $\mathbb{R}^{q_1}$. 

If, on the other hand, the network is deep, the vector $\boldsymbol{z}^{(1)} (\boldsymbol{x})$ is used as input in the next layer for computing new features and so on for the following layers. Let $h \in \mathbb{N}$ be the number of hidden layers (depth of network), and $q_k \in \mathbb{N}$, for $1\le k \le h$, be a sequence of integers  that indicates the dimension of each FCN layer (widths of layers). A deep FCN can be described as follows:

\begin{equation*}
\boldsymbol{x}  \mapsto  \boldsymbol{z}^{(h:1)} (\boldsymbol{x})= \left( \boldsymbol{z}^{(h)} \circ \cdots \circ \boldsymbol{z}^{(1)} \right) (\boldsymbol{x}) \in \mathbb{R}^{q_h},
\end{equation*}

where the vector functions $\boldsymbol{z}^{(k)}:\mathbb{R}^{q_{k-1}} \to \mathbb{R}^{q_{k}}$  have the same 
structure, and 
$W^{(k)} = (\boldsymbol{w}^{(k)}_j)_{1\leq j \leq q_k} \in \mathbb{R}^{q_k \times q_{k-1}}$,  $\boldsymbol{w}^{(k)}_0 \in \mathbb{R}^{q_k}$, 
for $1 \leq k \leq h$ are the network weights. In the case of deep NN, the output layer uses the features extracted by the last hidden layer  $\boldsymbol{z}^{(h:1)}(\boldsymbol{x})$ instead of those $\boldsymbol{z}^{(1)}(\boldsymbol{x})$. 

Finally, we mention that if the inputs to the FCN have a sequential structure, one way of processing these is to apply the same FCN to each input in turn, producing learned features for each entry in the sequence. This is called a point-wise neural network in \cite{vaswani2017attention} and a time-distributed network in the {\tt Keras} library.  

\subsection{Embedding Layer}

An embedding layer is designed to acquire a low-dimensional representation of categorical variable levels. Let $q_\mathcal{L} \in \mathbb{N}$ denote the hyperparameter determining the size of the embedding. The categorical variable levels are transformed into a real-valued $q_\mathcal{L}$-dimensional space, and the coordinates of each level in this new space serve as learned parameters of the neural network, requiring training; see \cite{guo2016entity} who introduced this technique for deep learning models.

The distances between levels in this learned space reflect the similarity of levels concerning the target variable: closely related levels exhibit small Euclidean distances, while significantly different categories display larger distances.

Formally, let $\mathcal{L} = {l_1, l_2,\dots, l_{n_\mathcal{L}} }$ represent the set of categories for the qualitative variable, and $n_\mathcal{L}$ denote its cardinality. The embedding layer functions as a mapping
\begin{align*}
\bz_\mathcal{L} : \mathcal{L} \to \mathbb{R}^{q_\mathcal{L} }.
\end{align*}

The total number of embedding weights to be learned during training is $n_\mathcal{L}q_\mathcal{L}$.

\subsection{Attention Layer}

An attention layer is a component in neural network architectures that implements an attention mechanism, which, in some sense ``focuses" on certain aspects of the input data that are deemed to be relevant for the problem at hand. These mechanisms allow the flexibility and performance of models to be enhanced and have produced excellent results, especially in tasks involving sequences like natural language processing, as well as within actuarial tasks, see, for example, \cite{kuo2021embeddings}. The main idea of attention mechanisms is their ability to enable deep neural networks to reweight the significance of input data entering the model dynamically.

Several attention mechanisms have been proposed in the literature. We focus on the most popular form of attention, which is the \emph{scaled dot-product} attention proposed in \cite{vaswani2017attention} for as a component of the Transformer model proposed there.

Let $Q \in \mathbb{R}^{q\times d}$ be a matrix of query vectors, $K\in \mathbb{R}^{q\times d}$ be a matrix of key vectors,  $V\in \mathbb{R}^{q\times d}$ is a matrix of value vectors. The scaled dot-product attention mechanism is a mapping: 

\begin{equation*}\label{FCN layer}
	A: \mathbb{R}^{(q \times d)\times (q \times d) \times (q \times d)} \to \mathbb{R}^{(q \times d)}, \quad \quad (Q,K,V) \mapsto A  = \tt{attn}( Q,K,V ). 
\end{equation*}

The attention mechanism is applied to the matrix $V$,  and the resulting output is calculated as a weighted sum of its elements. 
These attention coefficients depend on the matrices $Q$ and $K$. They undergo scalar-dot multiplication first, followed by the application of the softmax function to normalize the scores.
 Formally, the attention mapping has the following structure:
 
\begin{equation*}\label{attention}
 A = \text{softmax}(B) V = \tt{softmax}\bigg(\frac{Q K^{\top}}{\sqrt{d}} \bigg) V
\end{equation*}

where $d \in [0, +\infty)$ is a scalar coefficient, and the matrix of the scores $B^{*}$  is derived from the matrix $B$:
\[
B^* = \tt{softmax}(B) \quad \tt{where} \quad b^*_{i,j} = \frac{\exp(b_{i,j})}{\sum_{k=1}^{q} \exp(b_{i,k})} \in (0,1).
\]

$d$ can be set equal to the dimension of the value vectors in the attention calculation, or can alternatively be a learned parameter. The scalar dot attention mechanism is computationally efficient, mainly because it does not require recursive computation and is thus easily implemented on Graphics Processing Units (GPU), and has been widely adopted in Transformer-based models due to its simplicity and effectiveness in capturing relationships between elements in a sequence.

We provide a simple intuition for the learned attention scores in $B^{*}$, which are multiplied by the value vectors in $V$. Each row of the new matrix  $B^*V$ is comprised of a weighted average of the vectors in $V$, where the weights - adding to unity - determine the ``importance" of each row vector $i$ of $V$.

\subsection{Other network layers}

Here we briefly other neural network layers that were tested when designing the attention network that is the main result presented in Section \ref{sec:core}. 

The network components presented to this point process inputs to the network without any reference to previous inputs to the network. Recurrent Neural Networks (RNNs) modify neural networks to maintain an internal state that is calibrated from previous inputs of the network. An example of an RNN is the Long Short Term Memory (LSTM) network of \cite{hochreiter1997long}, which updates its internal state using sub-networks that either ``update" the internal state or ``forget" information that has already been learned.

Convolutional Neural Networks (CNNs) differ from FCNs by connecting the units of this network only to a small patch of the inputs, whereas the units in FCNs are connected to all of the inputs. In a CNN, the same network weights are applied to each patch.

Finally, a Transformer model \cite{vaswani2017attention} builds on the self-attention layer in two main ways: first, instead of applying self-attention once, multi-head attention is used to derive several versions of the attention matrix $A$ which are then compressed into a single matrix, and second, a point-wise neural network is used to process the compressed outputs from the multi-head attention.

We refer to \cite{goodfellow2016deep} and \cite{Wuthrich2021} for more detail on these network layers.

\subsection{Network Calibration}

The network's performance hinges on appropriately calibrating the weights in different layers, denoted as $w^{(k)}_{l,j}$. In the case of Fully Connected Neural (FCN) layers, these weights manifest as matrices $W^{(k)}$ and a bias term $\boldsymbol{w}_0^{(k)}$ for each $k$ ranging from 1 to $m$. Meanwhile, for Embedding (EN) layers, the weights correspond to the coordinates of levels in the new embedding space, represented as $\boldsymbol{z}_l (l)$ for all $l$ in $\mathcal{L}$.

The training process involves unconstrained optimization, where a suitable loss function $L(w^{(k)}_{l,j}, \cdot)$ is chosen, and the objective is to find its minimum. The Neural Network (NN) training employs the Back-Propagation (BP) algorithm, wherein weight updates are determined by the gradient of the loss function. The iterative adjustment of weights aims to minimize the error between the network outputs and reference values. The training complexity increases with the number of layers and units per layer in the network architecture; these are hyperparameters that should be suitably chosen. Indeed, a too deep NN  would lead to \emph{overfitting} producing a model unable to generalise to new data points, or, alternatively, lead to the \emph{vanishing gradient} problem, which prevents the BP algorithm from updating the weights successfully.
One remedy for the \emph{overfitting} problem is the application of regularization methods such as dropout.
Dropout \cite{srivastava2014dropout} is a \emph{stochastic} technique that ignores, i.e., sets to zero, some randomly chosen units during the network fitting. This is generally achieved by multiplying the output of the different layers by independent realizations of a Bernoulli random variable with parameter $p \in [0,1]$. Mathematically, the introduction of the dropout in a FCN layer, for example, in the $k$-th  layer, induces the following structure: 
\begin{eqnarray*}
r^{(k)}_j & \sim & Bernoulli(p)\\
 \dot{\boldsymbol{z}}^{(k-1)} (\boldsymbol{x}) & = & \boldsymbol{r}^{(k)}*  \boldsymbol{z}^{(k-1)} (\boldsymbol{x})\\
\boldsymbol{z}^{(k)} (\boldsymbol{x}) &=& \phi\left(\boldsymbol{w}^{k)} _0 +  W^{(k)} \dot{\boldsymbol{z}}^{(k-1)} (\boldsymbol{x})\right), \\
\end{eqnarray*}
where $*$ denotes the element-wise product and $\boldsymbol{r}^{(k)}$ is a vector of independent Bernoulli random variables, each of which has a probability $p$ of being $1$. This mechanism leads to zero the value of some elements and encourages more robust learning.

For a comprehensive understanding of neural networks and back-propagation, a detailed discussion can be found in \cite{goodfellow2016deep}.

\section{DeepYC - a neural network model for multiple yield curve modeling and forecasting}
\label{sec:core}

This section formally outlines the proposed deep learning-based yield curve model, which we call the DeepYC model. We have developed an architecture based on an attention mechanism, enabling efficient processing of a (multivariate) time series of yield time series, implicit information selection, and formulation of accurate predictions. Our goal is to derive precise point forecasts and effectively measure the uncertainty of the future yield curve. To achieve this, we have designed a model that allows us to jointly estimate a measure of the central tendency of the distribution of future spot rates (mean or median) and two quantiles at a specified tail level, forming confidence intervals for the predictions. 
Denoting as $\boldsymbol{y}_t^{(i)} = ({y}_t^{(i)}(\tau))_{\tau \in \mathcal{M}}\in \mathbb{R}^{M}$ the vector (yield curve) containing the spot rates for different maturities related to the curve family $i$, at time $t$. 

Choosing a look-back period $L \in \mathbb{N}$ and a confidence level $\alpha \in (0,1)$, we design a model that takes as input the the matrix of  yield curves related to the previous  $L$ dates denoted as  $Y^{{(i)}}_{{t}-L,{t}} = \big(\by^{{(i)}}_{{t-l}} \big)_{0\leq l \leq L}\in \R^{(L+1)\times M}$ and the label related to the curve family $i \in \mathcal{I}$, and produces three output vector $\boldsymbol{\widehat{y}}^{(i)}_{\alpha/2,t}, \boldsymbol{\widehat{y}}^{(i)}_{t}, \boldsymbol{\widehat{y}}^{(i)}_{1-\alpha/2,t}  \in \mathbb{R}^{M}$. More specifically, $\boldsymbol{\widehat{y}}^{(i)}_{\alpha/2,t}$ is the vector of quantiles at levels of confidence $\alpha/2$, $\boldsymbol{\widehat{y}}^{(i)}_{t}$ is the vector of a such measure of the central tendency of the distribution (in what follows, we consider both the mean or the median), and $\boldsymbol{\widehat{y}}^{(i)}_{1-\alpha/2,t} $ is the vector of quantiles at level $(1-\alpha/2)$. The quantile statistics will be calibrated to quantify the uncertainty in future yields.  
We desire to learn the mapping: 
\begin{align*}
f: \R^{(L+1) \times M} \times \mathcal{I} &\to \R^{M} \times \R^{M}  \times \R^{M}\\
      \big(Y^{(i)}_{{t}-L,t}, {i}\big) &\mapsto \big(\widehat{\by}^{(i)}_{\alpha/2, t+1}, \widehat{\by}^{(i)}_{t+1}, \widehat{\by}^{(i)}_{1-\alpha/2, t+1} \big) =f\left(Y^{{(i)}}_{{t}-L,{t}}, {i}\right).
\end{align*}
We approximate $f(\cdot)$ with a DNN that combines embedding,  FCN and attention layers. More specifically, we process the label $i$ using an embedding layer of size $q_{\mathcal{I}}\in \mathbb{N}$. It is a mapping with the following structure:
\[
	\boldsymbol{e}_{\cal I} : \mathcal{I} \to \mathbb{R}^{q_{\cal I}}, \quad \quad i \mapsto  \boldsymbol{e}_{\cal I}(i) =\left( e_{{\cal I},1}(i), e_{{\cal I},2} (i), \dots , e_{{\cal I},q_{\cal I}}(i) \right)^\top, 
\]
where $ \boldsymbol{e}_{\cal I} (i)$ is a vector that encoded the information related to the family of curves $i$. It can be seen as a new representation  of $i$ in a new $q_{\cal I}$-dimensional real-valued space that is optimal with respect to the response variable. 
On the other hand, we process the matrix of the past yield curves  with three time-distributed  FCN layers aiming to derive respectively the query, key and value vectors for input into the attention component.  The time-distributed mechanism consists of applying the same transformation (the same layer) to each row of the matrix $Y_{t-L,t}^{(i)}$ containing the yield curves of the different dates. We apply three FCN layers that can be formalised as: 
\begin{equation*}\label{FCN layer}
\boldsymbol{z}^{(j)} : \mathbb{R}^{M} \to \mathbb{R}^{q_A}, \quad \quad \by^{{(i)}}_{{t}} \mapsto  \boldsymbol{z}^{(j)} (\by^{{(i)}}_{{t}})
\end{equation*}
where $q_A$ is the number of units, and $j \in \{Q, K,V\}$.They produce the following matrices:
\begin{eqnarray*}
Q_t^{(i)}= \big(\boldsymbol{q}^{{(i)}}_{{t-l}}\big)_{0\leq l \leq L}\in \R^{(L+1)\times q_A}, &\quad &\boldsymbol{q}^{{(i)}}_{{t-l}}=\boldsymbol{z}^{(Q)} (\by^{{(i)}}_{{t-l}}) =  \phi^{(Q)}  \big(\boldsymbol{w}_0^{(Q)} + W^{(Q)}\by_{t-l}^{(i)}\big)\in \mathbb{R}^{q_A}\\
K_t^{(i)}= \big(\boldsymbol{k}^{{(i)}}_{{t-l}} \big)_{0\leq l \leq L}\in \R^{(L+1)\times q_A}, &\quad &\boldsymbol{k}^{{(i)}}_{{t-l}} =  \boldsymbol{z}^{(K)} (\by^{{(i)}}_{{t-l}}) = \phi^{(K)}  \big(\boldsymbol{w}_0^{(K)} + W^{(K)}\by_{t-l}^{(i)}\big)\in \mathbb{R}^{q_A}\\
V_t^{(i)}= \big(\boldsymbol{v}^{{(i)}}_{{t-l}} \big)_{0\leq l \leq L}\in \R^{(L+1)\times q_A}, &\quad &\boldsymbol{v}^{{(i)}}_{{t-l}} =\boldsymbol{z}^{(V)} (\by^{{(i)}}_{{t-l}}) = \phi^{(V)}  \big(\boldsymbol{w}_0^{(V)} + W^{(V)}\by_{t-l}^{(i)}\big)\in \mathbb{R}^{q_A}, 
\end{eqnarray*}
where $\boldsymbol{w}_0^{(j)} \in \mathbb{R}^{q_A}, W^{(j)}\in \mathbb{R}^{^{q_A} \times M}$ are network parameters, and $\phi^{(j)}: \mathbb{R} \mapsto \mathbb{R}$ are activation functions $, j \in \{Q, K, V\}$.
The three matrices  $Q_t^{(i)}, K_t^{(i)}, V_t^{(i)}$ are processed by an attention layer that combine them according to the following transformation:
\[
X_t^{(i)} = \text{softmax}\bigg( \frac{Q_t^{(i)}(K_t^{(i)})^\top}{\sqrt{d_k}}\bigg)V_t^{(i)}\in \R^{(L+1)\times {q_A}}
\] 
The output of this layer is arranged in a vector $\boldsymbol{x}_t^{(i)} = vec({X}_t^{(i)})$ that can be interpreted as a set of features that summarizes the information contained in the matrix $Y_{t-L,t}^{(i)}$. We remark that other deep learning models could be used to derive the vector of features from the matrix of the past data; here, we use the attention mechanism, having found that this produces the best out-of-sample forecasting performance.

Finally, we apply three different $|\mathcal{M}|$-dimensional FCN layers to the set of features $\boldsymbol{x}_t^{(i)}$ for deriving the predictions related to the three vectors of output statistics of interest. The first layer aims to derive the lower quantiles of the yields, the second one aims to compute the mean (or the median), while the third one is related to the upper quantiles. Each unit of these layers is dedicated to a specific maturity and we have in each layer as many units as the maturities considered.  Before inputting the outputs of the attention layer $\boldsymbol{x}_t^{(i)}$ to these FCNs, we apply dropout to $\boldsymbol{x}_t^{(i)}$ for regularization and reducing overfitting. In this setting, the predictions are obtained according to the following set of equations: 
\begin{align}\nonumber
 r^{(\text{att})}_j & \sim Bernoulli(p^{(\text{att})}_1)\\ \nonumber
\dot{\bx}_t^{(i)} & = \boldsymbol{r}^{(\text{att})} * \bx_t^{(i)}\\
\widehat{\by}^{(i)}_{ t+1}&=g \bigg( \boldsymbol{b}_c+U_c \boldsymbol{e}^{(i)} +W_c \dot{\bx}_t^{(i)} \bigg)\label{expected}\\
      \widehat{\by}^{(i)}_{\alpha/2, t+1}&=\widehat{\by}^{(i)}_{ t+1}-\phi_+ \bigg(\boldsymbol{b}_{lb}+U_{lb} \boldsymbol{e}(i) +W_{lb} \dot{\bx}_t^{(i)} \bigg)\label{lowerb}    \\
          \widehat{\by}^{(i)}_{1-\alpha/2, t+1}&=\widehat{\by}^{(i)}_{ t+1}+\phi_+ \bigg( \boldsymbol{b}_{ub}+U_{ub} \boldsymbol{e}(i) +W_{ub}\dot{\bx}_t^{(i)} \bigg)\label{uppebr}
\end{align}
where $p^{(\text{att})}_1 \in [0,1]$ is the dropout rate,  $g: \mathbb{R} \to \mathbb{R} $  and $\phi_{+} : \mathbb{R} \to ]0, +\infty)$ are strictly monotone functions, and $\boldsymbol{b}_j, U_j, W_j$,  $j \in \{c,lb, ub\}$ are network parameters.

Looking at this set of equations, some remarks can be made:

\begin{itemize}
\item[(1)] The model presents some connections with the affine models\footnote{The term ``affine term structure model'' is used in different ways by the literature, we refer to the definition given in Chapter 12 of \cite{piazzesi2010affine}.} discussed in  \cite{piazzesi2010affine}. 
Considering a single maturity $\tau$, equation \eqref{expected} can be formulated as follows: 
\[
g^{(-1)}\big(\widehat{\by}^{(i)}_{ t+1} (\tau)\big)= {b}_c+ \left\langle\boldsymbol{u}_{c, \tau}, \boldsymbol{e}^{(i)} \right\rangle+ \left\langle\boldsymbol{w}_{c, \tau}, \dot{\bx}_t^{(i)}\right\rangle.
\]
Indeed, it has the constant-plus-linear structure and depends on the vector of variables $\bx_t^{(i)}$ derived by the past observed data. Furthermore, the following additional arguments can be provided:
\begin{itemize}
    \item ${b}_c$ can be interpreted as a sort of global intercept; 
    \item $\left\langle\boldsymbol{u}_{c, \tau},\boldsymbol{e}^{(i)} \right\rangle$ is an intercept correction related to the  curve family $i$ and maturity $\tau$;
    \item  $\boldsymbol{w}_{c, \tau}$ is the vector of maturity-specific weights associated with $\bx_t^{(i)}$. These coefficients are shared among all the curve families since these do not depend on $i$.
\end{itemize}
\item[(2)]  The formulation we propose avoids potential quantile crossing, which refers to the scenario in which the estimated quantiles of a probability distribution do not respect the expected order. Inaccurate or inconsistent quantile estimates have the potential to affect the reliability and interpretability of predictions from the model.
Indeed,  the use of the activation function $\phi_+(\cdot)$ that assumes only positive values ensures that: 
\[
\widehat{\by}^{(i)}_{\alpha/2, t+1} < \widehat{\by}^{(i)}_{ t+1} < \widehat{\by}^{(i)}_{\-\alpha/2, t+1}.
\]

\item[(3)] Rewriting equations \eqref{lowerb} for a single maturity, we have:
\[
\phi^{-1} \bigg(\widehat{y}^{(i)}_{t+1}(\tau)-\widehat{y}^{(i)}_{\alpha/2, t+1}(\tau) \bigg) =  {b}_{lb}+ \left\langle\boldsymbol{u}_{lb, \tau}, \boldsymbol{e}^{(i)} \right\rangle+ \left\langle\boldsymbol{w}_{lb, \tau}, \dot{\bx}_t^{(i)}\right\rangle
\]
emphasizing that we model, on the $\phi^{(-1)}$ scale, the difference between the central measure and lower quantile at a given maturity $\tau$ is an affine model. Similar comments can be made for the difference between the upper quantile and the central measure. 
\end{itemize}

\begin{figure}
	\centering
		\includegraphics[scale = 1]{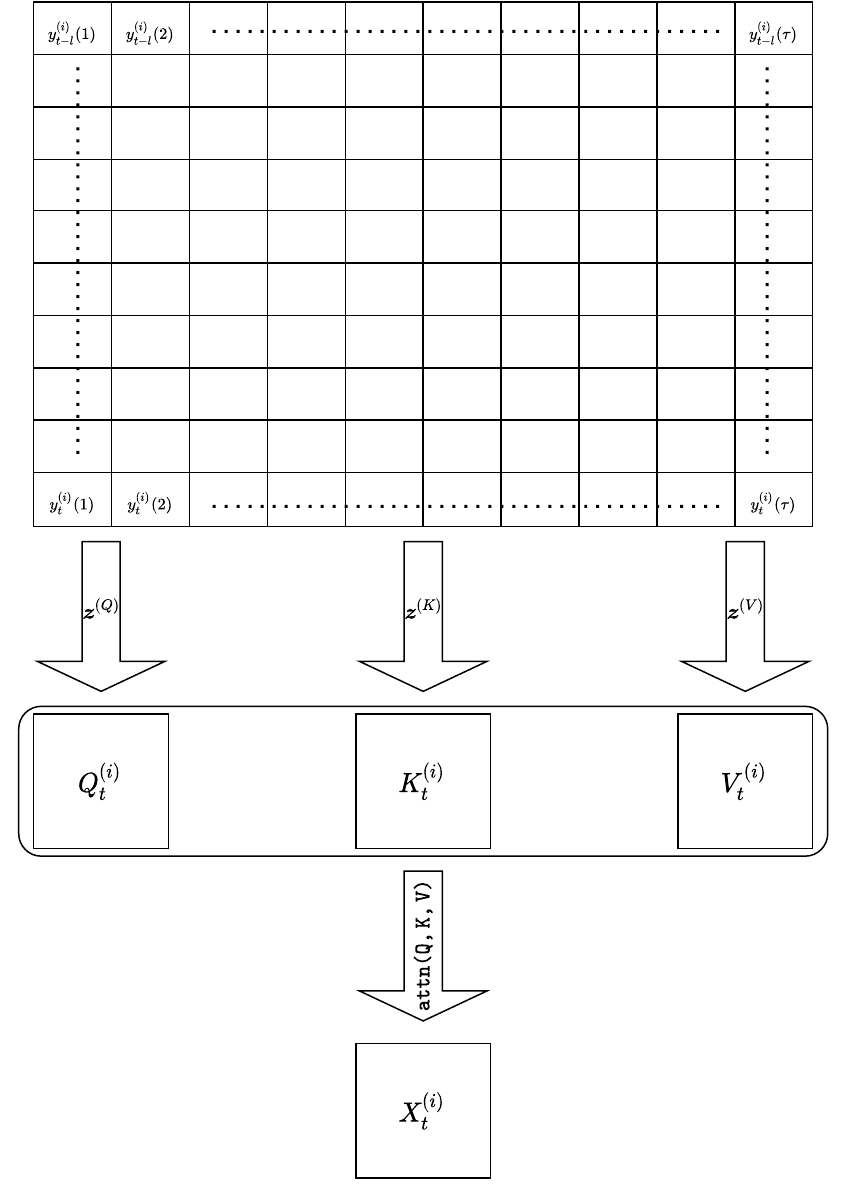}
\caption{Diagram of the feature processing components of the DeepYC model. A matrix of spot rates is processed by three FCN layers in a time-distributed manner, to derive the key, query and value matrices which are then input into a self-attention operation.}
	\label{fig:deepyc}
\end{figure}

\begin{figure}
	\centering
		\includegraphics[scale = 1]{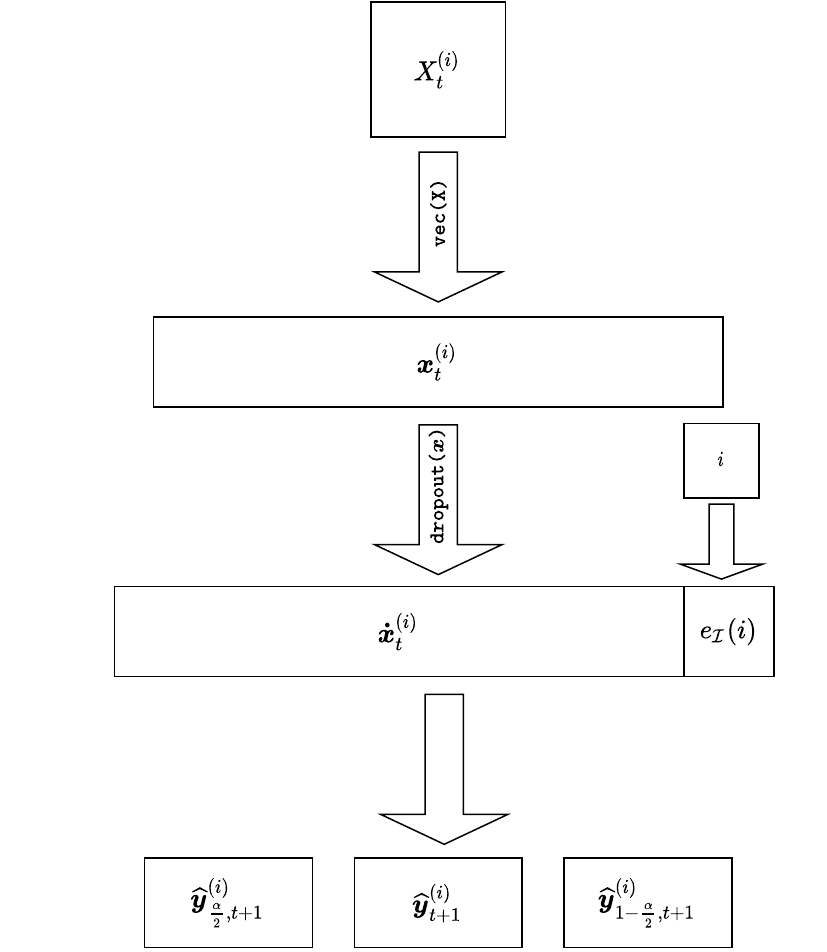}
\caption{Diagram of the output components of the DeepYC model. The matrix of features produced by the first part of the model are flattened into a vector and then dropout is applied. We add a categorical embedding to this vector, and, finally, then the best-estimate and quantile predictions are produced.}
	\label{fig:deepyc_out}
\end{figure}

We illustrate the DeepYC model in Figures \ref{fig:deepyc} and \ref{fig:deepyc_out}.

The calibration of the multi-output network is carried out accordingly, with a loss function specifically designed for our aim. It is the sum of three components:

\begin{align}
\label{eq:full_loss}\nonumber
\mathcal{L}_{\alpha, \gamma}(\boldsymbol{\theta}) &= \mathcal{L}_{\alpha/2}^{(1)}(\boldsymbol{\theta}) + \mathcal{L}_{\gamma}^{(2)}(\boldsymbol{\theta}) + \mathcal{L}_{1-\alpha/2}^{(3)}(\boldsymbol{\theta}) \\
&= \sum_{i,t, \tau} \ell_{\alpha/2} (y^{(i)}_t(\tau) - \hat{y}^{(i)}_{\alpha/2,t}(\tau)) + \sum_{i,t, \tau} h_{\gamma} (y^{(i)}_t(\tau) - \hat{y}^{(i)}_{t}(\tau)) + \sum_{i,t, \tau} \ell_{1-\alpha/2} (y^{(i)}_t(\tau) - \hat{y}^{(i)}_{1-\alpha/2, t}(\tau))
\end{align}

where $\ell_{\alpha}(u), \alpha \in [0,1]$ is the pinball function: 
\[
    \ell_{\alpha}(u) = \begin{cases}
(1-\alpha)|u| \quad u \leq 0 \\
\alpha |u| \quad \quad \quad \ \  u >0,
\end{cases}\]
and $h_{\gamma}(u), \gamma\in \{1,2\}$ is:
\[
    h_{\gamma}(u) = \begin{cases}
|u| \quad \gamma = 1\\
u^2 \quad\, \gamma = 2,
\end{cases}\]

We emphasize that the first term of the loss function is the pinball function with parameter $\alpha/2$ associated with the estimation of the lower quantile. The second term represents a generic function linked to the estimation of the central tendency, while the last term is the pinball function with parameters $\alpha/2$ associated with the estimation of the upper quantile.

Regarding the function $h_{\gamma}(\cdot)$, it is noteworthy that setting $\gamma=2$ means that we use the Mean Squared Error (MSE), indicating that we are modeling the mean of the yields; alternatively, setting $\gamma=1$, $h_{\gamma}$ is the the Mean Absolute Error (MAE), signifying that we are modeling the median.

\section{Numerical Experiments}
\label{sec:numerical}

We present some numerical experiments conducted on data provided by the European Insurance and Occupational Pensions Authority (EIOPA). The authority publishes risk-free interest rate term structures derived from government bonds of various countries; these are published on a monthly basis as spot curves, which are the default option for use in the Solvency II regime for discount rates. We define $\mathcal{I}$ as the set of all available countries and $\mathcal{M} = \{\tau \in \mathbb{N}: \tau \leq 150\}$. Our sample data covers the period from December 2015 to December 2021, and Figure \ref{fig:data} provides a graphical representation of the EIOPA data.

Selecting an observation time $t_0$, we partition the full dataset into two parts. The first, containing yields before time $t_0$ (referred to as the \emph{learning sample}), is used for model calibration. The second, encompassing data after time $t_0$ (referred to as the \emph{testing sample}), is employed to evaluate the out-of-sample accuracy of the models.  Interval forecasts are constructed by considering the case in which we desire a coverage probability $\alpha = 0.95$. 

To benchmark our model, we consider the Dynamic NS proposed in \cite{diebold2006forecasting} and its related Svensson (NSS) extension. We examine both cases where the latent factors follow individual AR(1) models and the case of a single multivariate VAR(1) model, denoted respectively as NS\_AR (NSS\_AR) and NS\_VAR (NSS\_VAR).

Since our focus is on measuring forecasting accuracy in terms of both point and interval forecasts, we employ several metrics to compare the models. Concerning point forecast accuracy, we use the global (i.e., evaluated for all countries in the dataset) Mean Squared Error (MSE) and the Mean Absolute Error (MAE) defined as follows:

\[
    \text{MSE} = \frac{1}{n} \sum_{i \in \mathcal{I}} \sum_{t \in \mathcal{T}} \sum_{\tau \in \mathcal{M}} (y^{(i)}_{t}(\tau)  - \hat{y}^{(i)}_{t}(\tau) )^2, 
\]
\[
    \text{MAE} = \frac{1}{n} \sum_{i \in \mathcal{I}} \sum_{t \in \mathcal{T}} \sum_{\tau \in \mathcal{M}} |y^{(i)}_{t}(\tau)  - \hat{y}^{(i)}_{t}(\tau) |, 
\]
where $n \in \N$ is the number of instances. We emphasize that MSE  is based on the $l_2$-norm of the errors and penalizes a larger deviation from the observed values with respect to the MAE which is based on the $l_1$-norm of the errors. For measuring the interval prediction accuracy, we consider the Prediction Interval Coverage Probability (PICP):
\begin{equation}
    \label{eq:global_PICP}
    \text{PICP}= \frac{1}{n} \sum_{i \in \mathcal{I}} \sum_{t \in \mathcal{T}} \sum_{\tau \in \mathcal{M}}  \mathbbm{1}_{\{ y^{(i)}_{t}(\tau)  \ \in \ [\hat{y}^{(i)}_{t, LB}(\tau), \ \hat{y}^{(i)}_{t, UB}(\tau)]\}}  
\end{equation}
Furthermore, we also analyse the Mean Prediction Interval Width (MPIW) to take into account the width of the confidence interval:
\begin{equation}
\label{eq:MPIW}
\text{MPIW} = \frac{1}{n} \sum_{i \in \mathcal{I}} \sum_{t \in \mathcal{T}} \sum_{\tau \in \mathcal{M}} \big( \hat{y}^{(i)}_{t, UB}(\tau)-\hat{y}^{(i)}_{t, LB}(\tau) \big).
\end{equation}

Intuitively, a good model should provide a PICP close to the value of $\alpha$, indicating that the model is well calibrated, while having as small an MPIW, as possible. Table \ref{tab:model_metrics_extended} presents the performance related to the four measures considered for the NS and NSS benchmark models.

\begin{table}
  \centering
  \begin{tabular}{lcccc}
    \hline
    Model   & MSE    & MAE    & PICP   & MPIW  \\
    \hline
    NS\_AR   & 0.7433 & 0.4496 & 0.9984 & 0.0540 \\
    NS\_VAR  & 0.4977 & 0.3492 & 0.7288 & 0.0080 \\
    NSS\_AR  & 0.5379 & 0.3709 & 0.9987 & 0.4253 \\
    NSS\_VAR & 0.4626 & 0.3226 & 0.7462 & 0.0307 \\
    \hline
  \end{tabular}
  \caption{Performance of the NS and NSS models in terms of MSE, MAE, PICP and MPIW; The MSE values are scaled by a factor of $10^5$, while the MAE values are scaled by a factor of $10^2$.}
  \label{tab:model_metrics_extended}
\end{table}
Regarding the accuracy of point forecasts, it is worth noting that employing a VAR(1) model instead of independent AR(1) models for modeling the latent factors dynamics enhances the performance of both the NS and NSS models. This improvement is evident in terms of both Mean Squared Error (MSE) and Mean Absolute Error (MAE). Furthermore, it is noteworthy that NSS models consistently demonstrate superior accuracy compared to NS models.
Turning our attention to interval forecasts, it becomes apparent that models incorporating AR processes tend to exhibit over coverage, as indicated by excessively high Prediction Interval Coverage Probability (PICP) and impractically large interval widths providing very limited information content. In contrast, versions based on VAR models for NS and NSS tend to yield more reasonable interval widths. However, given their relatively low coverage probability, they fall short of adequately capturing the uncertainty in future yields. In summary, we designate the NSS\_VAR model as the optimal choice, as it yields the lowest MSE and MAE, along with the highest PICP and a reasonable MPIW. Consequently, we will employ this model for subsequent comparisons throughout the remainder of the paper.

Now, we can focus on the YC\_ATT model. We examine two variants  distinguished by the gamma parameter in the loss function employed for calibration. The first variant is calibrated by setting $\gamma = 1$ (denoted as YC\_ATT$_{\gamma = 1}$) and, in that case, the second component of the loss function is the MAE. On the other hand, the second variant is calibrated with $\gamma = 2$ and the component of the loss related to the central tendency of the distribution then uses the MSE.
Regarding the other hyperparameters, we have configured the number of units in the dense layers comprising the attention component of the model as $q_A = 8$, while  the dropout rate is set  $p_1^{(att)} = 0.5$.
As previously indicated, the part  of the network architecture that processes the past yields can be constructed using various neural network blocks. The attention layer, discussed earlier, is just one of the available choices. We further explore the utilization of other well-established deep learning models that have demonstrated success in modeling sequential data. Specifically, we investigate other 3 variants:
\begin{itemize}
    \item YC\_LSTM: based on the Long Short Term Memory network (LSTM) of \cite{hochreiter1997long} that is a popular kind of RNN; 
    \item A simplification of the YC\_ATT  model, that removes the attention mechniasm and relies only on processing the yield curves using time-distirubted FCNs; since these are also called one-dimensional convolutional neural networks, we call this variant YC\_CONV; 
    \item YC\_TRANS which is a more complex Transformer based model \cite{vaswani2017attention}, which adds extra FCNs to the YC\_ATT  model. 
\end{itemize}
{In order to make the comparison fair, also for these variants we set the number of units in the layers equal to $8$. }
Furthermore, for each one of these architectures we respectively test the variants with $\gamma = 1$ and $\gamma = 2$. 

Table \ref{tab:average_ensemble} presents the performance metrics, including Mean Squared Error (MSE), Mean Absolute Error (MAE), Prediction Interval Coverage Probability (PICP), and Mean Prediction Interval Width (MPIW), for various deep learning models. 
 \begin{table}
\centering
\resizebox{\textwidth}{!}{
\begin{tabular}{l|cc|cc|cc|cc}
                         & \multicolumn{2}{c|}{MSE}          & \multicolumn{2}{c|}{MAE}          & \multicolumn{2}{c|}{PICP}         & \multicolumn{2}{c}{MPIW}          \\
Model                    & average         & ensemble        & average         & ensemble        & average         & ensemble        & average         & ensemble        \\ \hline
YC\_ATT$_{\gamma = 1}$   & \textbf{0.2947} & \textbf{0.2887} & \textbf{0.2667} & \textbf{0.2616} & \textbf{0.9154}          &0.9191 & 0.0106          & 0.0106          \\
YC\_ATT$_{\gamma = 2}$   & 0.3663          & 0.3638          & 0.3463          & 0.3451          & 0.8528          & 0.8573          & 0.0105          & 0.0105          \\
YC\_CONV$_{\gamma = 1}$  & 0.3778          & 0.3642          & 0.2975          & 0.2850          & {0.9035} &  \textbf{0.9235}          & 0.0115          & 0.115           \\
YC\_CONV$_{\gamma = 2}$  & 0.4258          & 0.4244          & 0.3890          & 0.3884          & 0.8509          & 0.8530          & 0.0110          & 0.0110          \\
YC\_LSTM$_{\gamma = 1}$  & 0.4272          & 0.4111          & 0.3164          & 0.2970          & 0.7757          & 0.8147          & 0.0093          & 0.0093          \\
YC\_LSTM$_{\gamma = 2}$  & 0.3898          & 0.3697          & 0.3352          & 0.3198          & 0.6911          & 0.7081          & \textbf{0.0084} & \textbf{0.0084} \\
YC\_TRAN$_{\gamma = 1}$  & 0.4308          & 0.4167          & 0.3313          & 0.3168          & 0.8371          & 0.8645          & 0.0113          & 0.0113          \\
YC\_TRANS$_{\gamma = 2}$ & 0.4232          & 0.4124          & 0.4042          & 0.3987          & 0.5771          & 0.5760          & 0.0091          & 0.0091          \\ \hline
\end{tabular}

}
\caption{Out-of-sample performance of the different deep learning models in terms of MSE, MAE, PICP and MPIW; the MSE values are scaled by a factor of $10^5$, while the MAE values are scaled by a factor of $10^2$. Bold indicates the smallest value, or, for the PICP, the value closest to $\alpha = 0.95$.}
\label{tab:average_ensemble}
\end{table}

To account for randomness in batch sampling and the random initial parameters for each network for the optimization, multiple training attempts were conducted, and the results represent the average performance over 1  training attempts.  The boxplots related to the 10 training attempts are shown in Figure \ref{fig:boxplot} (Appendix A). Additionally, ensemble predictions were generated by averaging the forecasts from the ten trained models. We remark that the average MPIW model on the different training attempts and the MPIW related to the ensemble predictions coincide. The proof of this statement is reported in the Appendix.

Several noteworthy findings emerge. Firstly, the consistent adoption of ensemble mechanisms leads to superior performance compared to the average performance across individual models. This trend holds across all examined models and metrics, aligning with prevailing findings in the literature on predictive modeling with deep learning. Moreover, a notable trend is highlighted: models configured with $\gamma = 1$ consistently outperform their counterparts with $\gamma = 2$ in both point forecasts and interval forecast accuracy. This outcome can be attributed to the robust nature of MAE as a measure, providing increased resilience to outliers and contributing to a more stable model calibration. Suprisingly, this is even the case when evaluating the models using the MSE metric. Finally, among the deep learning models, the YC\_ATT model with $\gamma = 1$ demonstrates the best performance, suggesting that this architecture is particularly well-suited for the regression task at hand. Also notable is that the ensemble predictions of the  YC\_ATT model with $\gamma = 1$ are almost the best calibrated, as measured by the PICP metric, while the bounds are relatively narrow, as measured by the MPIW metric. Nonetheless, the best performance on the PICP metric on a standalone basis is the YC\_CONV with $\gamma = 1$. Regarding ensemble predictions, it is noteworthy that the average MPIW  model across various training attempts coincides with the MPIW associated with ensemble predictions. The proof of  this assertion can be found in the Appendix. In summary, evaluating the models on all of the metrics, the YC\_ATT performs the best overall.

Figure \ref{fig:density} graphically compares the point and interval forecasts generated by the NSS\_VAR and YC\_ATT models for yield curves across various countries. 
 \begin{sidewaysfigure}
\includegraphics[width=\columnwidth]{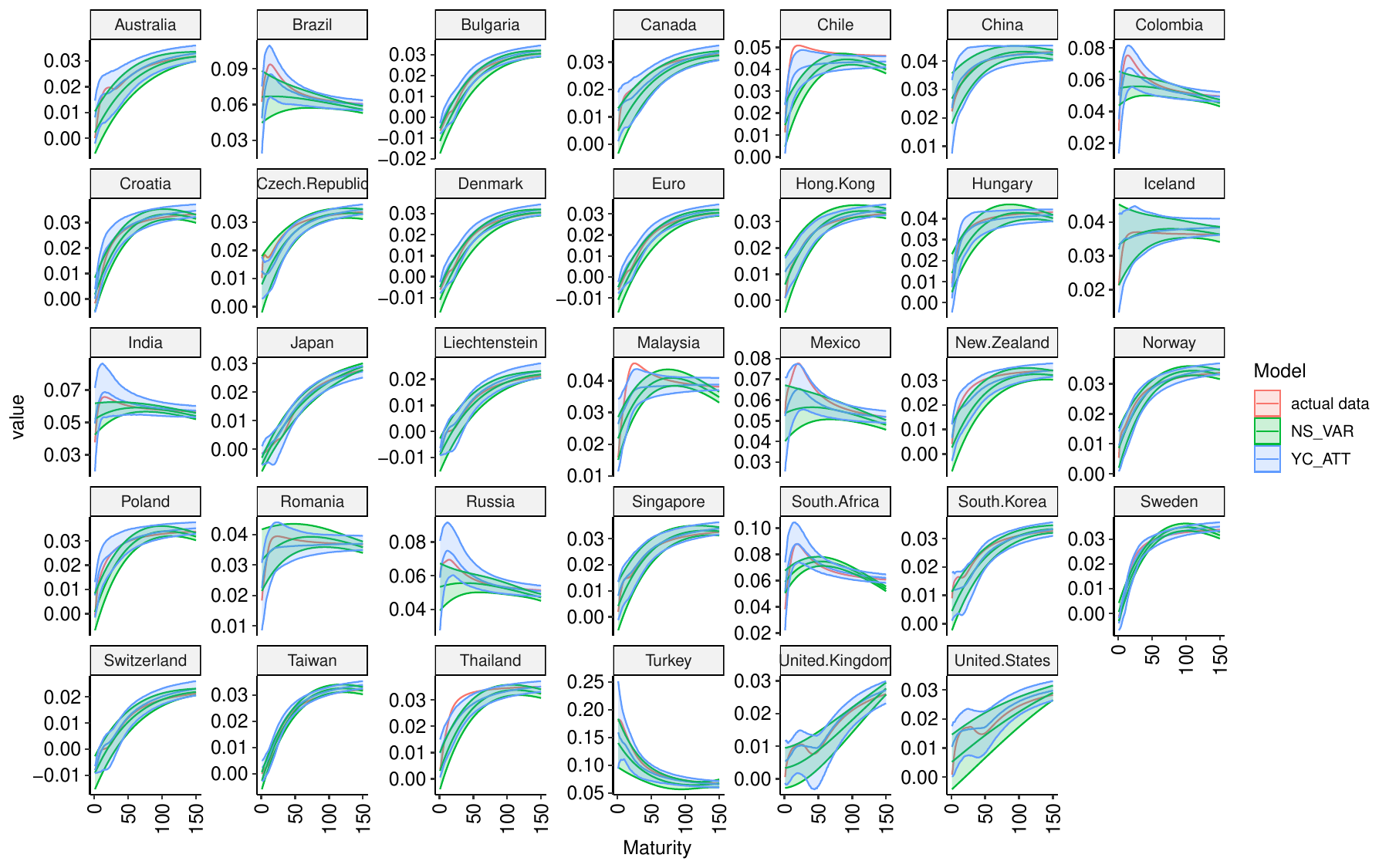}%
  \caption{Point and interval forecasts for EIOPA yield curves  generated by the NSS\_VAR and YC\_ATT models as of June 2021, the central date of the forecasting period. } 
    \label{fig:density}
\end{sidewaysfigure}
The figure refers to the central date within the forecasting horizon, specifically June 2021, as the forecasting period spans monthly observations from January to December 2021. The figure shows the actual observed yield curve, and an out-of-sample forecast of the best-estimate and quantiles using the previous 10 months of data; model parameters are those calibrated using data up to the end of 2020. 
Upon examination, it becomes evident that the NSS\_VAR model falls short in accurately depicting the shape of yield curves in certain countries. Notably, the realized yields deviate significantly from the projected forecasts and extend beyond the confidence intervals. This discrepancy is particularly pronounced in the cases of Brazil, Colombia, Mexico, India, Russia, South Africa, and others.
In contrast, the YC\_ATT model exhibits more flexibility and demonstrates the ability of effectively capturing the uncertainties associated with future yields.

\begin{figure}
	\centering
		\includegraphics[scale = 0.5]{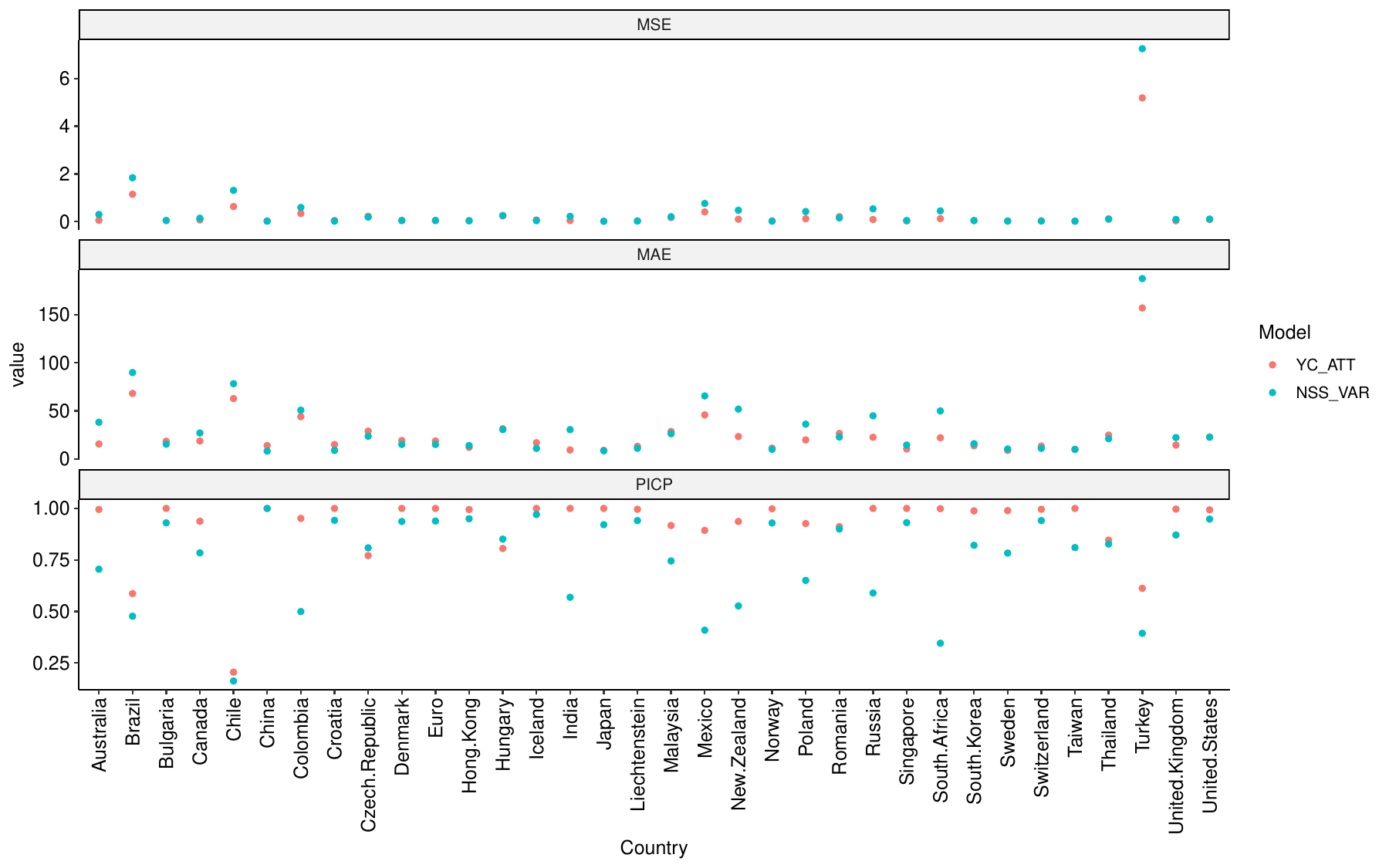}
\caption{MSE, MAE, and PICP obtained by the YC\_ATT  and NSS\_VAR models in the different countries. }
	\label{fig:res_country}
\end{figure}

Figure \ref{fig:res_country} offers a more in-depth analysis of the performance exhibited by the YC\_ATT and NSS\_VAR models across various countries of the EIOPA yield curves. It illustrates the MSE, MAE, and PICP produced by the different models for the different yield curve families.  From the standpoint of point forecasts, it is noted that in certain instances, the YC\_ATT and NSS\_VAR models produce comparable results. However, for specific cases, the YC\_ATT model demonstrates a notable enhancement, particularly evident in developing countries such as Brazil, Chile, Malaysia, Mexico, South Africa, and Turkey. This finding suggests that yields shows that when the yield curve evolution can be adequately described by the NSS\_VAR model, our YC\_ATT tends to replicate the same point predictions. On the other hand, for the yield curve families for which the NSS\_VAR model model is not optimal, the YC\_ATT improve the results. 
In terms of interval forecasts, there is a noticeable improvement attributable to the YC\_ATT model, impacting all the EIOPA yield curves under consideration.


 \begin{sidewaysfigure}
\includegraphics[width=\columnwidth]{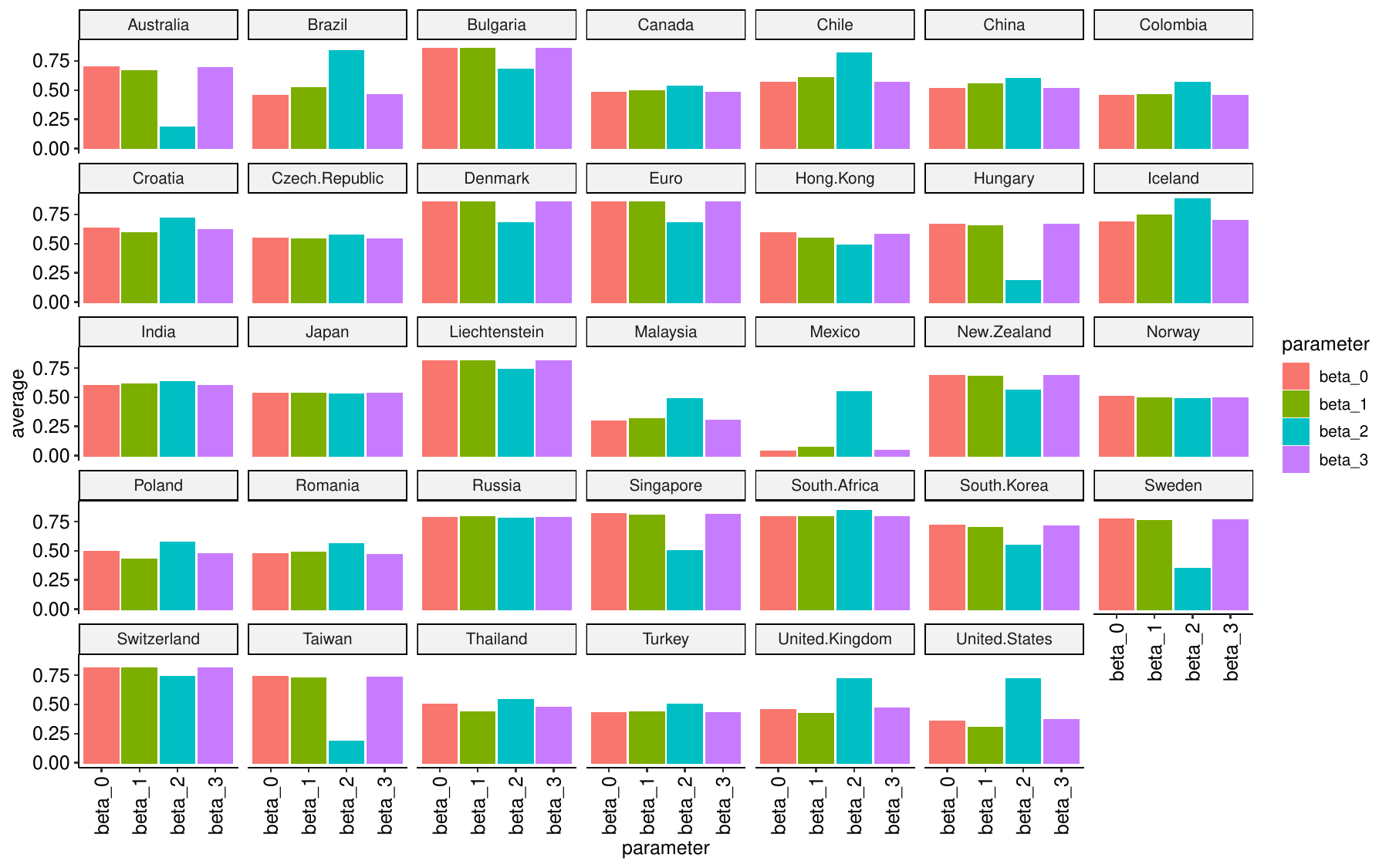}%
  \caption{Point and interval forecasts for the yield curves of various families are generated by the NSS\_VAR and YC\_ATT models for the EIOPA data in the central date of forecasting timeframe. } 
    \label{fig:density}
\end{sidewaysfigure}

To gain insights into the mechanism underlying the YC\_ATT model, we investigate the features that the model extracts from the input data and that are used by the output layers to derive key statistics. We consider   the feature vector $(\boldsymbol{e}(i), \boldsymbol{x}_t^{(i)}) \in \mathbb{R}^{q_{\mathcal{I}}+q_{A}\times M}$, and conduct Principal Component Analysis (PCA) to reduce the dimensionality. We extract the first four Principal Components (PCs) which explain the 97\% of the variability of $(\boldsymbol{e}(i), \boldsymbol{x}_t^{(i)})$ such that we have a mapping with the structure $\mathbb{R}^{q_{\mathcal{I}}+q_{A}\times M} \mapsto \mathbb{R}^4$.
To assess the similarity of information contained in the four PCs with the $\boldsymbol{\beta}_t^{(i)}$ factors of the NSS model, we compute the linear correlation between these two sets of features. Figure 4 visually represents the average absolute value of the Pearson linear correlation for each yield curve family. Notably, we observe that very high correlations are detected in some cases, while low correlations are obtained in the others. 
Examining this figure in conjunction with Figure 3, we note that in families exhibiting similar performances, such as Euro, Bulgaria, Denmark, and Iceland,  the PC components are high correlated with the NSS latent factors.
Conversely, instances of notable improvements by the YC\_ATT model, as seen in Mexico, Turkey, Malaysia, and Brazil, are accompanied by smaller correlations between the PC components from the output of the attention layer and the beta parameters of NSS.
In essence, this figure confirms that in cases where the yield curves follow a process adequately described by the NSS models, the YC\_ATT model replicates this by extracting variables highly correlated with beta. However, when this is not the case, and the yields present more complex patterns, our attention model derives features that deviate from the NNS model, resulting in better outcomes.






\section{Extensions and variants of the YC\_ATT model}
\label{sec:extensions}
In this section, we explore potential extensions and variants of the YC\_ATT, considering some modifications that aim to enhance the modeling of yield curves and their associated uncertainties.

\subsection{Deep Ensemble}
\begin{sidewaysfigure}
\includegraphics[width=\columnwidth]{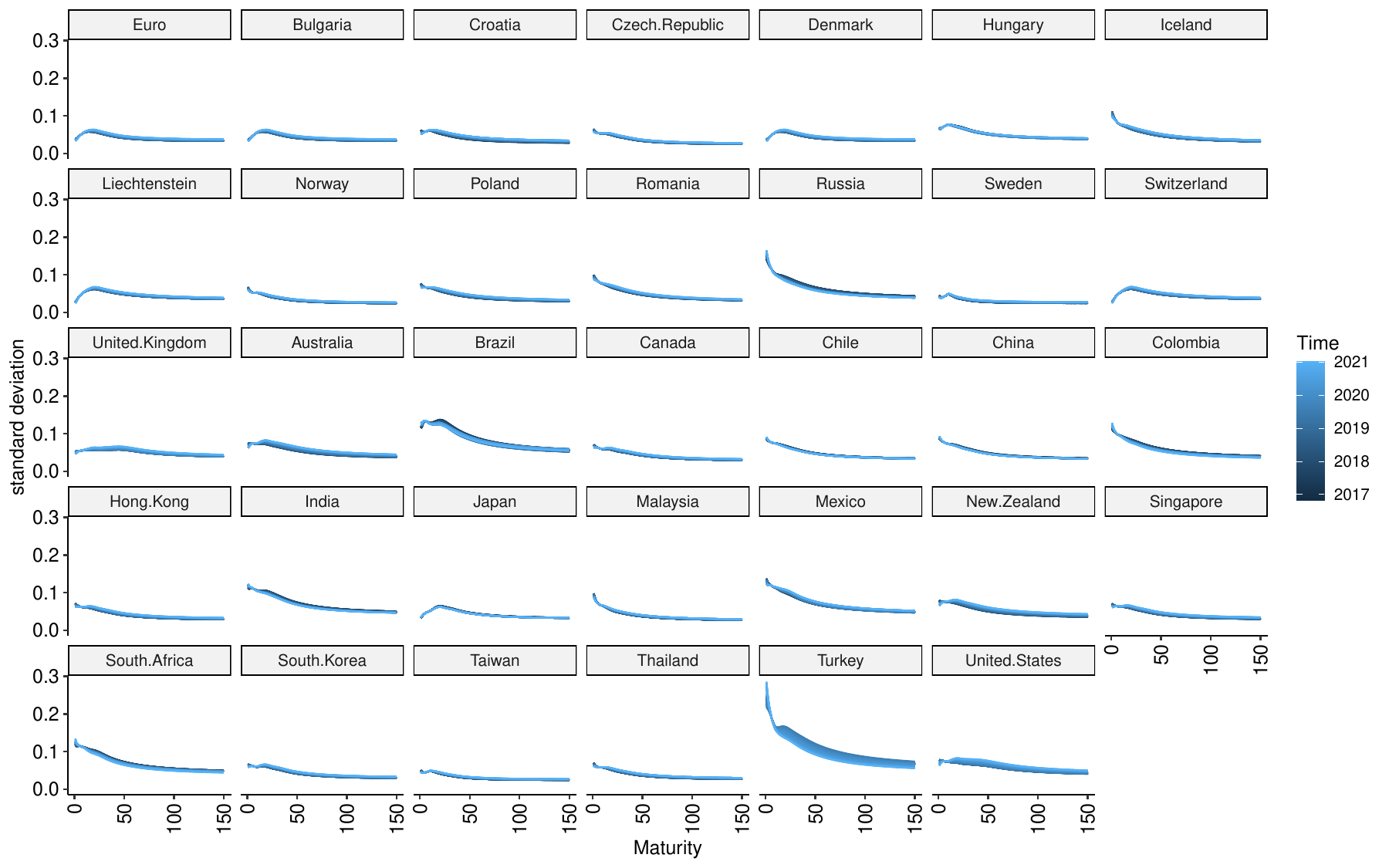}%
  \caption{$(\hat{\sigma}_t^{(i)}(\tau))_{\tau \in \mathcal{M}}$  estimates associated to the yields related to the different countries. } 
    \label{fig:variance_estimates}
\end{sidewaysfigure}

An alternative approach for modeling uncertainty in future yields is the Deep Ensemble (DE) method discussed in \cite{lakshminarayanan2017simple}. In contrast to the quantile regression-based YC\_ATT, this method relies on distributional assumptions for the response. The idea consists of formulating a heteroscedastic Gaussian regression model that provides joint estimates for both the mean and variance of the yields, denoted as $(y_{t}^{(i)}(\tau),(\sigma_{t}^{(i)}(\tau))^2)$. 
This technique not only facilitates the extraction of additional insights into future yields but also accommodates heteroscedasticity in the modeling process; this is different from the networks calibrated in the previous section which are equivalent to assuming that the responses follow a homoscedastic Laplace or Gaussian distribution for choices of $\gamma \in {0,1}$ respectively. Furthermore, confidence intervals can be derived using these estimates. In this vein, we design a network architecture with two output layers that produces predictions of the yield curves and their related variances. We refer to this model as YC\_ATT\_DE.
As discussed in \cite{nix1994estimating}, model calibration of the DE model can be performed by minimizing the following loss function:
\[
\mathcal{L}(\boldsymbol{\theta}) = \sum_{t,i,\tau} \bigg[ \frac{y_t^{(i)}(\tau) -\hat{y}_t^{(i)}(\tau) }{(\sigma_t^{(i)} (\tau))^2} + \frac{\log \big((\sigma_t^{(i)} (\tau))^2\big) }{2}\bigg].
\]
The first component represents the MSE between the prediction and the actual yields, scaled by the variance. Meanwhile, the second term acts as a penalization factor for observations with notably high estimated variances. 
We calibrate the YC\_ATT\_DE model on the EIOPA data in the same setting used above. Figure \ref{fig:variance_estimates} shows the  standard deviation  estimates $(\hat{\sigma}_t^{(i)}(\tau))_{\tau \in \mathcal{M}}$ associated to the yield curves for the different countries obtained through the YC\_ATT\_DE model. 

We notice that larger standard deviations are detected for the yields corresponding to short time to maturity in contrast to the yields associated with longer maturities. This observation aligns with intuition, as shorter-term yields are more susceptible to market fluctuations, rendering them more volatile. Additionally, we note that standard deviation estimates are notably higher for specific members of the EIOPA family of yield curves, specifically those linked to Turkey, Russia, Brazil, and Mexico. This finding is plausible in light of the economic instability experienced in these countries.

\begin{figure}
	\centering
		\includegraphics[scale = 0.6]{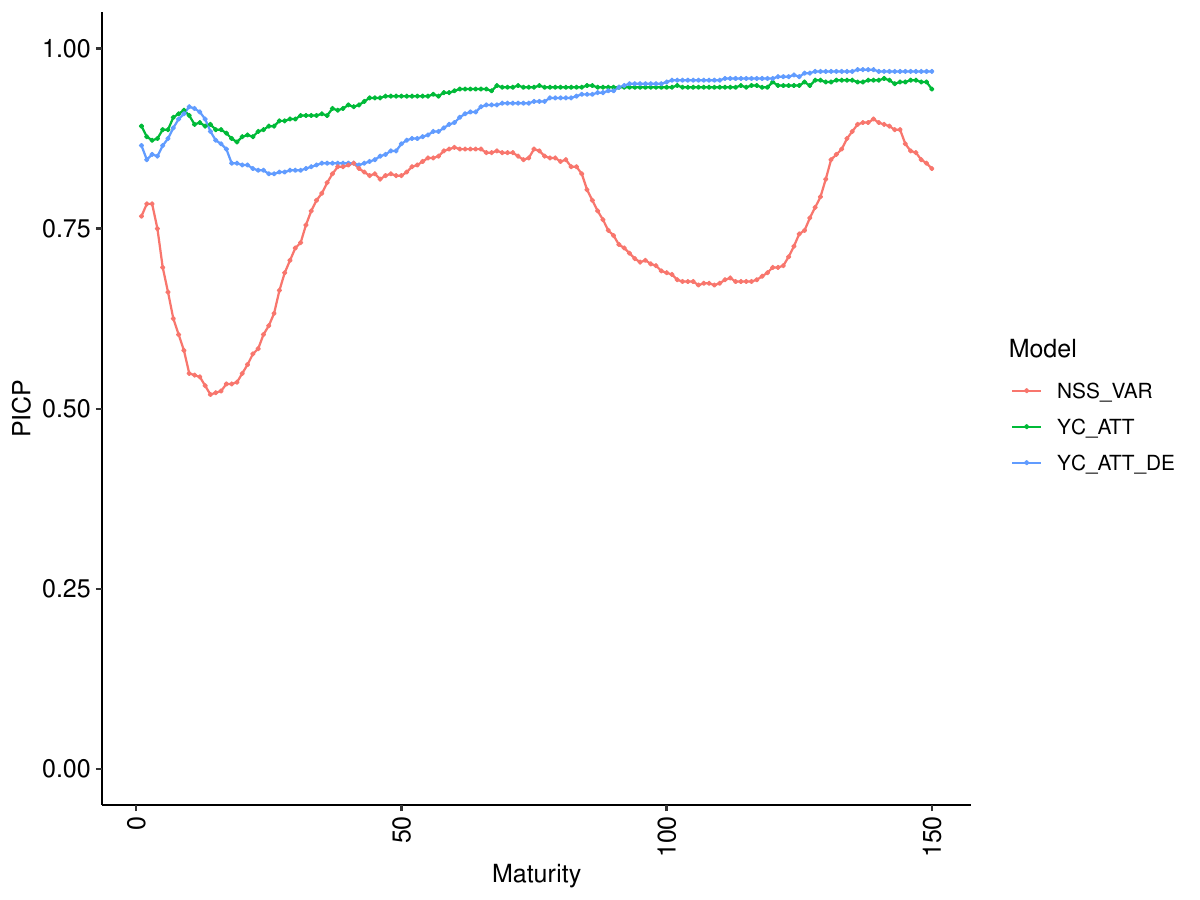}
\caption{ PICP of the NSS\_VAR, YC\_ATT, and YC\_ATT\_DE models for different time-to-maturities. }
    	\label{fig:PICP_DE}
\end{figure}

Figure \ref{fig:PICP_DE} illustrates the PICP of the YC\_ATT, YC\_ATT\_DE, and NSS\_VAR models across various time-to-maturities. Notably, the NN-based models exhibit significantly superior performance compared to the NSS\_VAR model, confirming once again that the NSS model is not sufficiently flexible to capture uncertainty in certain yield curve families. On the other hand, YC\_ATT and YC\_ATT\_DE emerge as more promising candidates to address this task, producing higher PICP for all the maturities considered. Furthermore, upon comparing YC\_ATT and YC\_ATT\_DE, we observe that the former tends to excel for short time-to-maturity, while the latter yields higher PICPs for longer times to maturity. This finding suggests that the Gaussian distribution assumption appears to be more suitable for yields with long maturities,  since the yields related to short maturity are more susceptible to market fluctuations and may be affected by some asymmetry and fat tails.

We note that the forecast term structures of standard deviations from the YC\_ATT\_DE are a nice by-product of this method, and can be used for other quantitative risk management applications.

\subsection{Transfer Learning}
The calibrated YC\_ATT models may also provide some benefits when used on smaller datasets through the mechanism of transfer learning. Transfer learning allows for leveraging knowledge gained from solving one task and applying it to improve the performance of a different but related task. In other words, we take a model trained on one task (the source task) is repurposed or fine-tuned for a different but related task (the target task).

For this particular application, we aim to exploit a model with experience acquired in modeling and forecasting EIOPA yield curves to construct forecasting models for different families of yield curves. Transfer learning is of particular interest when a dataset of experience is available, that is too small to calibrate reliable models on. For example, with the recent implementation of IFRS 17, companies will produce portfolio specific illiquidity-adjusted yield curves. It is likely that these curves comprise too small a dataset to model; in this case transfer learning can be used.

To explore this idea, we collected a new, smaller dataset of US spot curves relating to assets with different rating levels. A visual representation of this supplementary data is presented in Figure \ref{fig:lstm} in the appendix.  In this context, the set of yield curve families is defined as $\dot{\mathcal{I}} = \{\tt{AAA}, \tt{AA}, \tt{A}, \tt{BBB}, \tt{BB}, \tt{B}\}$. The set of maturities is represented by \\ $\dot{\mathcal{M}} = \{0.25, 0.5, 1, 2, 3, 4, 5, 6, 7, 8, 9, 10, 15, 20, 25, 30\}$, denoting $\dot{M} = |\dot{\mathcal{M}}|$ its cardinality,  and the time span $\dot{\mathcal{T}}$ covers monthly observations from January 2015 to October 2021. Importantly, we here have different inputs to the networks, both in terms of the number of maturities and the categorical input $i$. We divided the dataset into a \emph{learning sample} and a \emph{testing sample}, conducting a forecasting exercise for the most recent 12 months of experience in this dataset.

 As noted, the two groups of curves present different number of maturities  ($\dot{M} \neq M$), and directly applying the YC\_ATT model to the new data becomes unfeasible. To address this issue, we equip YC\_ATT model with an additional layer designed to align the US credit curves with the same dimension as the EIOPA curves. This adjustment enables us to  process them using pre-calibrated attention layer of the YC\_ATT model, facilitating the extraction of the relevant features $\boldsymbol{x}_t^{(i)}$.
Denoting as $\tilde{Y}^{(i)}_{t-L, t} \in \mathbb{R}^{(L+1) \times \dot{M}}$  the matrix of past yield curves related to the $L+1$ previous date, we apply the (learned) mapping:
\[
\dot{\boldsymbol{z}}:\mathbb{R}^{(L+1) \times \dot{M}} \to \mathbb{R}^{(L+1)\times {M}}, \qquad\qquad  E_{t-L,t}^{(i)} = \dot{\boldsymbol{z}}( \dot{Y}^{(i)}_{t-L,t});
\]

this used an FCN layer. Furthermore, since we are now considering a different set of yield curve families, we also introduce a new embedding layer aimed to learn a $\mathbb{R}$-valued represetation of the elements in $\dot{\mathcal{I}}$ that is optimal with respect to the forecasting task. It is a mapping with the structure 
\[
	\dot{\boldsymbol{e}}_{\dot{\cal I}} : \dot{\mathcal{I}} \to \mathbb{R}^{q_{\dot{\cal I}}}.
\]
where $q_{\dot{\cal I}} \in \mathbb{N}$ is the hyperparameter defining the size of the embedding layer. 
{In this case, the three output layers related to the calculation of the lower quantiles, the best estimates, and the upper quantiles have size equal to $\dot{M}$.}

Letting $\dot{\boldsymbol{\theta}}$ be the vector of NN parameters of these two new layers, the calibration process is carried out by  minimizing the loss, as defined in Equation \ref{eq:full_loss}, which now also depends on the  parameters $\dot{\boldsymbol{\theta}}$. The objective is to learn an effective mapping that transforms the US credit curve data into the dimension of the EIOPA yield curve data in order to be processed by the pretrained attention layer of the YC\_ATT model and simultaneously train the new embedding layer. In essence, we optimise the model with respect to $\dot{\boldsymbol{\theta}}$ while keeping constant (or ``frozen")) the parameters  $\boldsymbol{\theta}^{(ATT)}$ related to the key, query and value FCNs and the attention layer:

\[
\argmin_{\dot{\boldsymbol{\theta}}} \mathcal{L}(\dot{\boldsymbol{\theta}}, \boldsymbol{\theta}^{(ATT)}).
\]

To benchmark our model with transfer learning - called YC\_transfer in the below - we present the comparison against the NS and NSS models. Since we are considering a different set of data, we now extend again the comparison to all four versions of the NS and NSS models with both $\text{AR}(1)$ and $\text{VAR}(1)$ parameter forecasts. We also include in the comparison the model $YC\_ATT$ that is directly trained on the US credit curve data.  For both NN-based models we also investigate the use of the ensemble mechanism. In this application, we focus on the case $\gamma = 1$, i.e., calibrating the best-estimates output of the model using the MAE.

\begin{sidewaysfigure}
\includegraphics[width=\columnwidth]{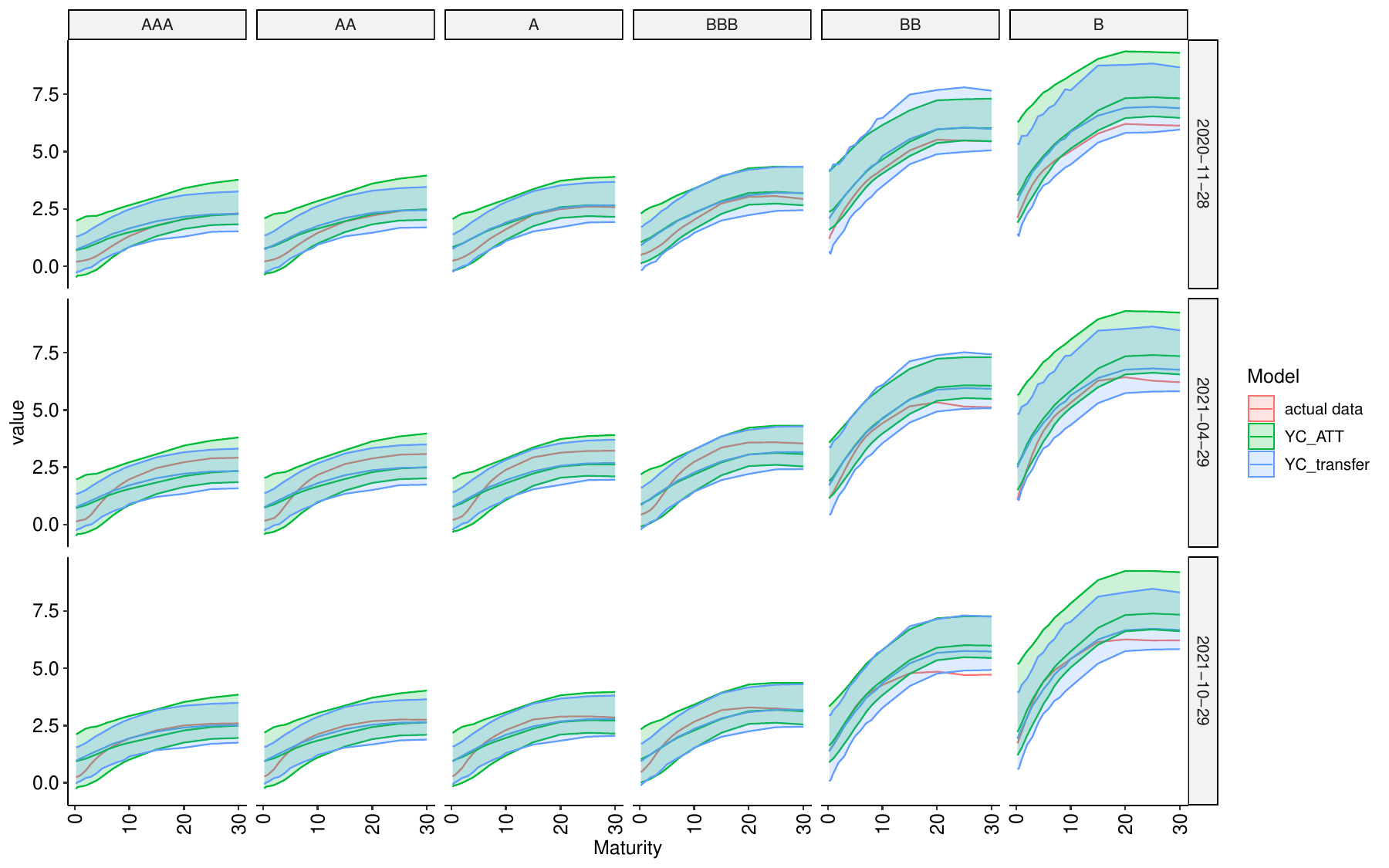}%
  \caption{Point and interval forecasts for the US credit curves  generated by the YC\_ATT  and YC\_transfer models for  the central, the middle and the final date of forecasting period. } 
    \label{fig:pi_bloom}
\end{sidewaysfigure}

Table \ref{tab:model_metrics_extended} presents the performance metrics for all the models across the four measures. 
\begin{table}
    \centering
    \begin{tabular}{l|ccccc}
        \hline
        Model & MSE & MAE & PICP & MPIW \\
        \hline
        NS\_AR & 0.4605 & 0.5279 & 0.5243 & 0.9221\\
        NS\_VAR & 0.3273 & 0.4412 & 0.43056 & 0.5796 \\
        NSS\_AR & 0.4833 & 0.5381 & 0.5512 & 1.0109 \\
        NSS\_VAR & 0.3416 & 0.4514 & 0.4323 & 0.8853 \\
        \hline
        YC\_ATT$_{\gamma = 1}$ 	& 0.3178&0.4691&{0.9003}&2.3189\\
         YC\_ATT$_{\gamma = 1}$ (ensemble) 	&0.3113&0.4640&0.9019&2.3189\\
              YC\_transfer$_{\gamma = 1}$ 	&0.3152&0.4622&\textbf{0.9285}&2.1109\\
         YC\_transfer$_{\gamma = 1}$ (ensemble) 	&\textbf{0.2262}&\textbf{0.3963}&0.9852&2.1109\\\hline
    \end{tabular}
    \caption{MSE, MAE, PICP and MPIW of the different models considered. Bold indicates the smallest value, or, for the PICP, the value closest to $\alpha = 0.95$.}
    \label{tab:my_table}
\end{table}
We note that NS\_VAR and NSS\_VAR models outperform their counterparts that are based on independent AR models, in terms of point forecasts. However, it is noteworthy that all four models exhibit poor performance in terms of PICP, indicating a limited ability to capture uncertainty surrounding future yields. When examining  NN-based models, we also note the ensemble mechanism consistently enhances the results of both the  YC\_ATT$_{\gamma = 1}$ and YC\_transfer$_{\gamma = 1}$ models. The most accurate outcomes are obtained with the YC\_transfer$_{\gamma = 1}$ (ensemble), which produces superior performance in terms of MSE, MAE, and PICP.
Figure \ref{fig:pi_bloom} illustrates the point and interval forecasts of the YC\_ATT$_{\gamma = 1}$ and the YC\_transfer$_{\gamma = 1}$ models  on three distinct dates: the starting date, the middle date, and the last date of the forecasting horizon.  
Notably, the width of the forecast interval expands as we transition from yield curves associated with high {\tt AAA} ratings to those with lower {\tt B} ratings. This evidence works for all three dates. This trend aligns with expectations, as greater uncertainty is logically anticipated in yields linked to lower-rated companies. Moreover, both models exhibit commendable performance in predicting yields for reliable ratings ({\tt BBB, A, AA, AAA}). The transfer model, in particular, demonstrates enhanced coverage for the lower-rated categories ({\tt B} and {\tt BB}), where more uncertainty is expected. A plausible explanation for this observation is that the transfer model adeptly captures tension by leveraging insights gained from EIOPA data, featuring yield curves marked by substantial volatility.

\section{Conclusions}
\label{sec:conclusions}

The accurate modeling of the yield curves is crucial in insurance and finance for several reasons, playing a fundamental role in risk management, investment decision-making, and asset-liabilities evaluation. This paper has advanced the field by developing deep learning models to describe the dynamics of multiple yield curves associated with diverse credit qualities or countries simultaneously. We have confirmed the intrinsic ability of these models to effectively describe large-dimensional time-series data and model non-linearity inherent in yield curve dynamics. Our study shows that these models outperform other well-known models such as the dynamic version of the Nelson and Siegel \cite{nelson1987parsimonious, diebold2006forecasting} and the related Svenson extension \cite{svensson1994estimating} in the multiple yield curve modeling and forecasting tasks. We performed several numerical experiments on the data provided by the European Insurance and Occupational Pensions Authority (EIOPA). Although different kinds of neural network architecture have been investigated, we discover the most promising results have been obtained by using the self-attention mechanism, which has proven successful in natural language processing.
Furthermore, we also investigate techniques for quantifying the uncertainty around predictions, a critical yet under-explored in the existing literature. We explored the use of nonparametric quantile regression and designed an architecture specifically designed to avoid quantile crossing issues. The numerical analysis of the performance, conducted in terms of prediction interval coverage probability and mean prediction interval width, shows the effectiveness of the proposed approach. We finally discuss two possible extensions and variants of the proposed model. The first one considers using the deep ensemble method for measuring uncertainty in forecasts. This approach, which requires the assumption of heteroscedastic Gaussian distributions of the response, appears to be promising, especially in describing the dynamics of yields with long time-to-maturities; the DE approach also produces a term structure of forecast volatility, which is a useful by-product.  The second extension uses a transfer learning mechanism that could be useful to exploit the experiences gained in modeling the EIOPA yield curves to improve the performance of a model related to a different set of yield curves. A numerical illustration of this approach is conducted considering the US credit curve data with different credit qualities (ratings). We show that this approach allows for improved forecasting performance, especially in terms of prediction intervals,  when the data we are considering are subject to more uncertainty. 

In future studies, we plan to explore the use of explainable deep learning techniques, in particular, the LocalGLMnet model introduced in \cite{richman2023localglmnet}, to model effectively the uncertainty inherent in future yield predictions in an intepretable manner. Initially designed for expected values, an extension to quantile modeling is interesting but also challenging due to the intricate issue of quantile crossing. Moreover, we would like to model jointly interest rates and other market variables using a similar model. Finally, our future research agenda extends to the investigation of other potential applications of attention and transfer models within the insurance domain. Specifically, we aim to explore their efficacy in non-life insurance fields, such as frequency-severity modeling.

\section*{Acknowledgements}
{The authors acknowledge the International Actuarial Association that financially supported this work through the ``Life Section Research Grant"  assigned to the project ``Multiple Yield Curve modeling and Forecasting using Deep Learning". The authors are grateful to the staff members at Old Mutual who provided the US credit curves.}

\bibliographystyle{acm}

\clearpage
\pagebreak

 \appendix

 \section{Appendix: Proof, data and plots}
Here we show that the average MPIW  across different training attempts coincides with the MPIW of the ensemble predictions obtained by averaging the predictions of the ten models.\\ 
We denote:
\begin{itemize}
    \item $y_{i,k}^{(l)}$ lower bound of the k-th model, in the i-th observation.
        \item $y_{i,k}^{(u)}$ upper bound of the k-th model, in the i-th observation.
        \item $i = 1, \dots, n$ is the index related to the different observations; 
        \item $k = 1, \dots, m$ is the index related to the different models.
\end{itemize}
The MPIW of the $k$-th model on the different data points is: 
\[
\text{MPIW}_k = \frac{1}{n}\sum_i^n (y^{(u)}_{i,k} -y^{(l)}_{i,k}).
\]
The average MPIW is: 
\[
\text{MPIW}_{average} = \frac{1}{m}\sum_{k=1}^m \text{MPIW}_k= \frac{1}{n * m} \sum_{k = 1}^m\sum_{i = 1}^n (y^{(u)}_{i,k} -y^{(l)}_{i,k})
\] 
the MPIW of the ensemble predictions is: 
\[
\text{MPIW}_{ensemble} = \frac{1}{n} \sum_{i=1}^n \Bigg(\frac{1}{m}\sum_{k=1}^m y^{(u)}_{i,k} -\frac{1}{m}\sum_{k=1}^m y^{(l)}_{i,k}\Bigg) = \frac{1}{n * m} \sum_{k = 1}^m\sum_{i = 1}^n (y^{(u)}_{i,k} -y^{(l)}_{i,k}).
\]
Then, we can conclude that: 
\[
\text{MPIW}_{average} = \text{MPIW}_{ensemble}.
\]

 \begin{sidewaysfigure}
 \includegraphics[width=\columnwidth]{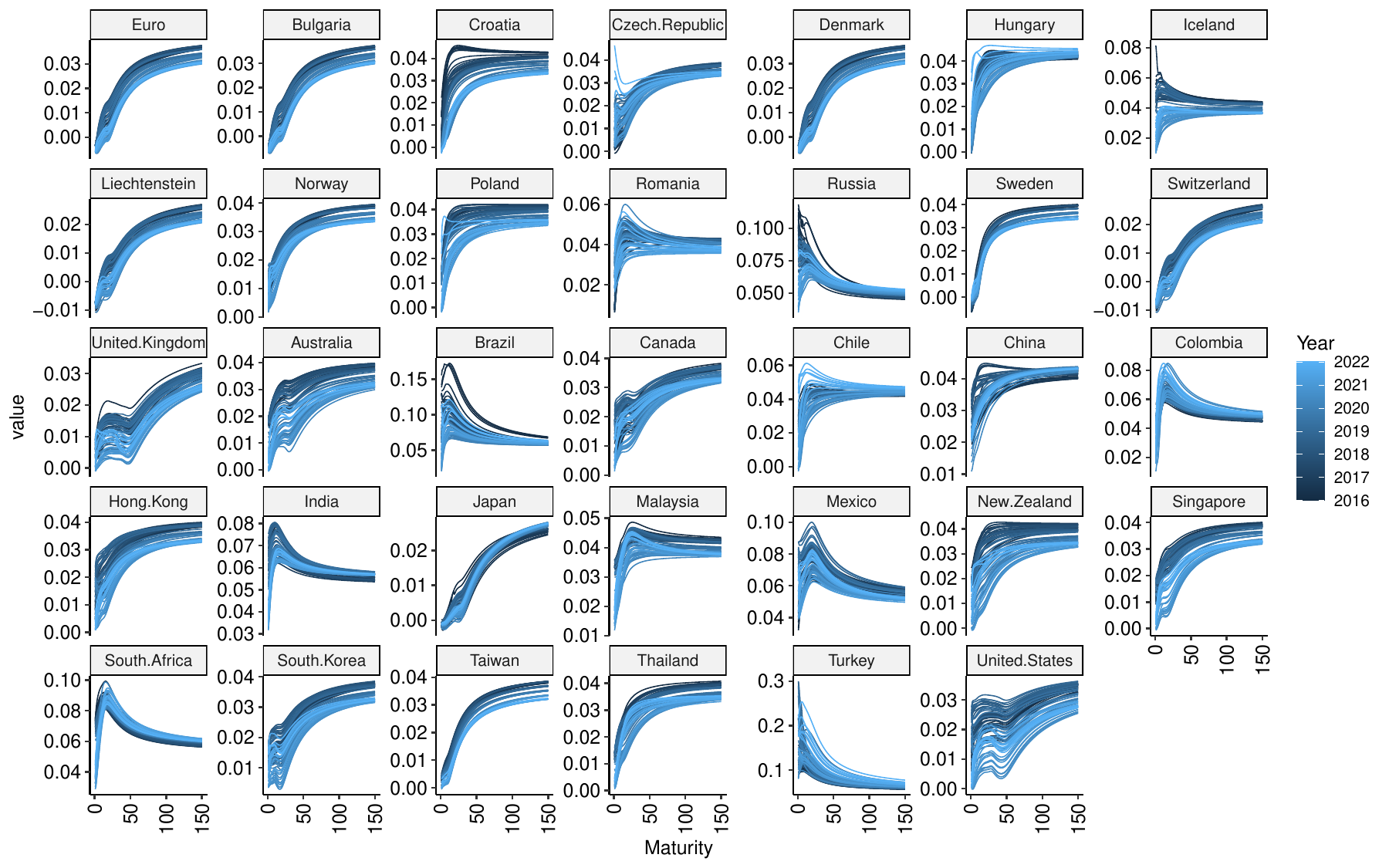}%
   \caption{Risk-free interest rate term structures derived from government bonds of different countries;observation period spans from December 2015 to December 2021.} 
     \label{fig:data}
 \end{sidewaysfigure}

\begin{figure}
	\centering
		\includegraphics[scale = 0.6]{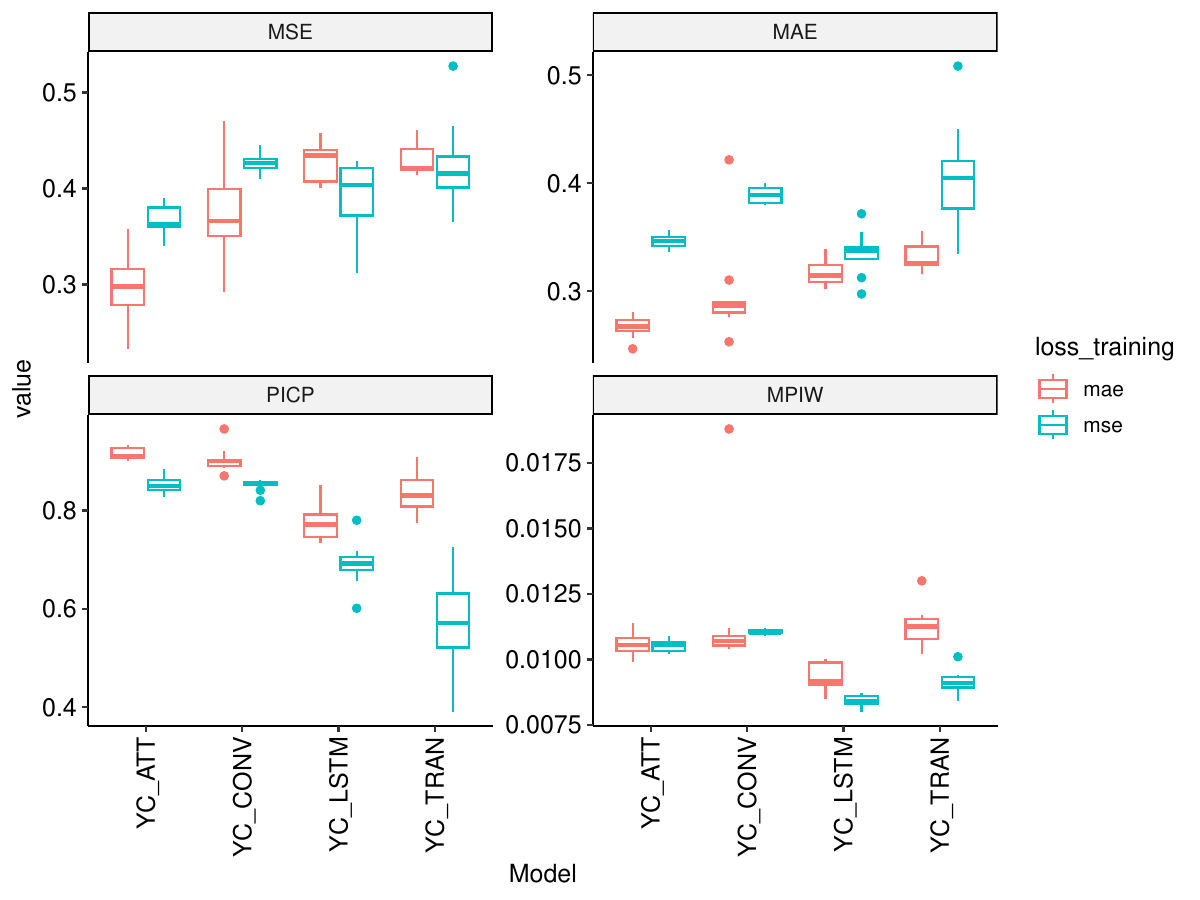}
\caption{Boxplot of the out-of-sample MSE, MAE, PICP and MPIW of the different models on ten runs; the MSE values are multiplied $10^5$, the MAE values are multiplied by $10^2$.  }
	\label{fig:boxplot}
\end{figure}

\begin{figure}
	\centering
		\includegraphics[scale = 0.65]{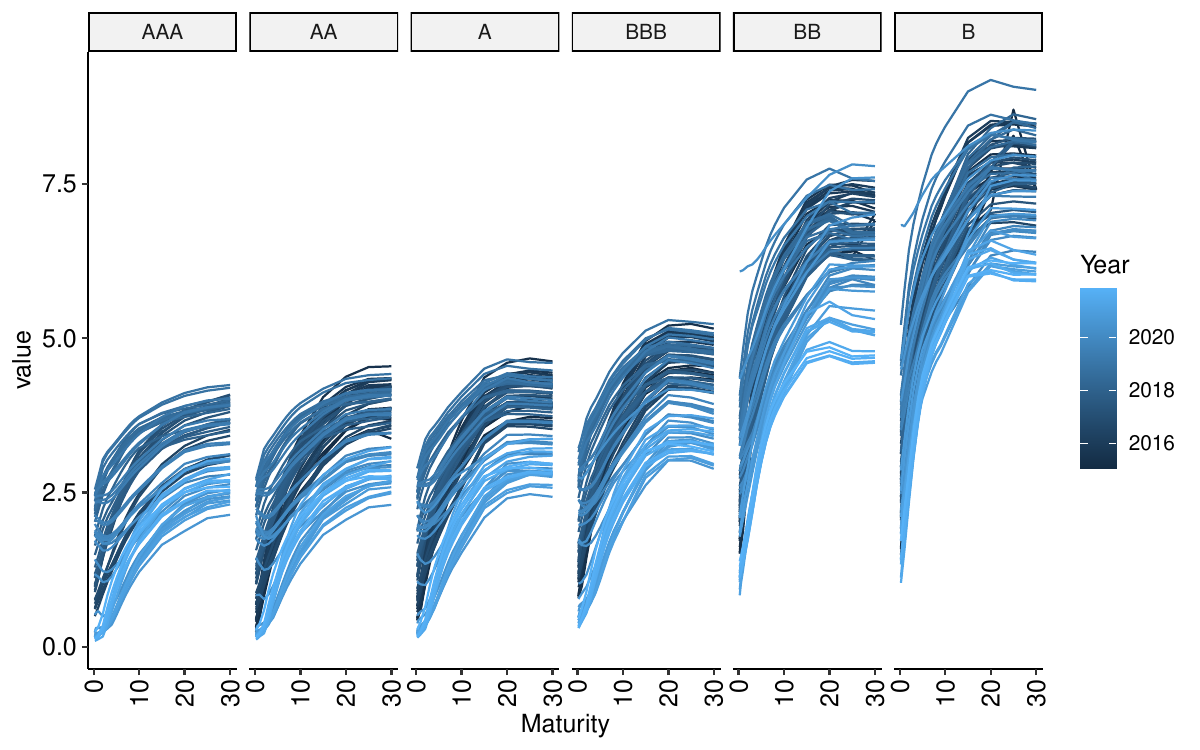}
\caption{US credit curves related to different rating qualities. }
	\label{fig:lstm}
\end{figure}

\end{document}